\documentclass{article}
\usepackage{iclr2027_conference,times}
\usepackage[T1]{fontenc}
\usepackage{amsmath,amssymb,mathtools}
\usepackage{graphicx,booktabs,array,microtype,placeins,needspace}
\usepackage{tabularx,longtable}
\usepackage[table]{xcolor}
\definecolor{rwHeader}{HTML}{F0F2F5}
\definecolor{rwGroupBlue}{HTML}{DDECF5}
\definecolor{rwGroupPurple}{HTML}{EAE5F2}
\definecolor{rwMethod}{HTML}{E7F2ED}
\definecolor{rwDelta}{HTML}{A52A2A}
\makeatletter
\newenvironment{rwtableside}{\def\@captype{table}\setlength{\abovecaptionskip}{0pt}\centering}{\par}
\makeatother
\usepackage{xurl}
\usepackage{hyperref,url}
\hypersetup{pdftitle={ReWorld-Track: A Recursive Event World Model for Language-Guided Multi-Camera Tracking},pdfauthor={Haoyang Wu, Shoudong Han, Chaoyue Li, Sijia Chen, Zhenyang Xie, Sihan Wang},pdfsubject={}}
\graphicspath{{figures/}}
\newcommand{\method}{ReWorld-Track}
\newcommand{\given}{\,\vert\,}

\newcommand{\Hist}{\mathcal H}
\newcommand{\Cand}{\mathcal Y}
\newcommand{\Camera}{\mathcal C}
\newcommand{\ind}{\mathbf 1}
\newcommand{\E}{\mathbb E}

\title{ReWorld-Track:\\A Recursive Event World Model for\\Language-Guided Multi-Camera Tracking}
\author{\parbox[t]{\dimexpr\textwidth-2\tabcolsep\relax}{\normalfont\small\raggedright
Haoyang Wu$^{1}$, Shoudong Han$^{1}$, Chaoyue Li$^{1}$, Sijia Chen$^{1}$, Zhenyang Xie$^{2}$, Sihan Wang$^{3}$\\
$^{1}$Huazhong University of Science and Technology; \quad
$^{2}$Jiangxi University of Water Resources and Electric Power; \quad
$^{3}$Zhongnan University of Economics and Law
}}
\iclrfinalcopy
\begin{document}
\maketitle
\lhead{Preprint}
\suppressfloats[t]
\begin{abstract}
Language-guided multi-camera tracking must preserve a target identity across unobserved gaps, where similar candidates and uncertain returns can make early associations unreliable. A wrong match can corrupt the history used to predict later observations and propagate identity errors across subsequent camera handoffs. We propose \textbf{ReWorld-Track}, a recursive event world model that carries association uncertainty into future predictions. Candidate matches and continued waiting define alternative target states, whose posterior probabilities are used to update a persistent recurrent belief. This representation preserves uncertainty about alternative trajectories through successive observations. This belief predicts the next camera, arrival time, and entry region, while appearance and language evidence guide association. By training across successive handoffs, the model learns to retain uncertainty that remains useful for later predictions and identity decisions. ReWorld-Track achieves HOTA scores of 65.19 on CityFlowV2 and 45.36 on MTMMC, with improved identity continuity across repeated handoffs. On MTMMC, its structured posterior update gains 0.50 HOTA points over a similarly sized generic updater and 0.94 points over fixed-moment soft association, raising next-camera accuracy from 86.03\% to 87.41\% and reducing median arrival-time error from 0.78 s to 0.71 s for subsequent target returns.
\end{abstract}

\section{Introduction}
\label{sec:intro}
Multi-camera tracking is an important computer vision task for understanding how people and vehicles move through an environment \citep{cityflow,mtmmc}. It links observations from different cameras into trajectories of the same physical targets. Road networks, campuses and large indoor spaces require this continuity because each camera covers only part of a route. Language-guided tracking lets a user specify the target through a description \citep{cityflownl,crmot}. We study the setting in which a query and an initial observation identify the person or vehicle to follow. The tracker must preserve this identity across views and wait when no observation supports a match. We call each transition between observed views a handoff.

\begin{figure}[t]
\centering
\includegraphics[width=\linewidth,trim=0 .25bp 0 3bp,clip]{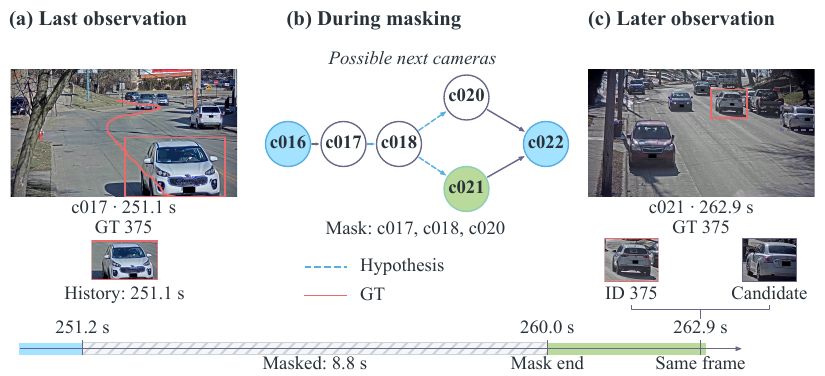}
\caption{Ambiguous identity continuation. (a) Last observation of GT375 in c017. (b) Camera hypotheses during the 8.8\,s mask. (c) Target and competitor in the same c021 frame. White nodes denote masked streams. Blue dashed edges show route hypotheses, and coral markings show ground truth (GT). The panels follow the same target through this sequence.}
\label{fig:problem}
\end{figure}

Handoffs are difficult because camera coverage is incomplete. A target may leave one view before entering another, creating a blind gap with no fresh image evidence. Its next camera and arrival time are uncertain. On return, a new viewpoint may hide a named attribute, while nearby objects resemble the target. Figure~\ref{fig:problem} illustrates the resulting ambiguity under controlled camera masking. After the last observation in panels~(a,b), several routes remain plausible. In panel~(c), the target and a competitor fit the same camera and timing forecast. Choosing the competitor changes both the identity and the history used for later handoffs. Natural gaps create the same lack of target observations while all camera streams remain available to detect a return.

Three challenges follow from this ambiguity. The tracker must predict through absence, distinguish similar returns, and preserve useful history despite uncertain matches. Motion and appearance support local tracking \citep{sort,deepsort}, while visual memory retains identity evidence across interruptions \citep{memotr,stmtrack,trackon}. Cross-camera forecasting estimates a future camera, arrival time and location \citep{trajectorytensors}. These capabilities help evaluate a return. The unresolved question is how the resulting association should change the state used for the following forecast. A single accepted candidate can erase plausible routes. Candidate probabilities alone leave the associated locations and motion histories unspecified. Continued absence also carries evidence about a route, depending on whether its camera is available. A useful update must carry these distinctions forward across successive handoffs.

Our central idea is to learn a predictive state from the uncertain association itself. We propose \method{}, a recursive event world model built around this idea. We use the term event world model to denote a task-specific predictive state model over future target observations, following the broader view of world models as learned predictive latent dynamics for future observations and downstream decisions \citep{worldmodels,planet,dreamer}. Unlike general-purpose or pixel-generative world models, ReWorld-Track predicts structured observation events: camera, arrival time and entry region. It recursively updates this predictive state after each association decision. It models the target's evolving observation process for subsequent tracking decisions.

Figure~\ref{fig:method} connects prediction, association and feedback. A recurrent model propagates the belief over elapsed time on a fixed camera graph. Its event forecast guides association together with appearance and query evidence. Association assigns probabilities to the candidates and a temporary-null alternative, representing an active target with no current match. A learned projection summarizes these weighted states for the next prediction, retaining route probabilities and motion uncertainty. During a wait, camera availability determines how continued non-observation changes route support. An unavailable stream supplies no negative visual evidence. Training spans successive handoffs, so later prediction errors shape the information retained from earlier ambiguous returns. This closes the loop between finding a likely return and maintaining the history needed to find the next one.

We evaluate whether this loop improves complete trajectories on CityFlowV2 vehicles and MTMMC pedestrians. ReWorld-Track reaches higher order tracking accuracy (HOTA) of 65.19 and 45.36, respectively. On MTMMC, it gains 1.24 HOTA points over adapted CRTracker and 0.50 over a comparably sized generic learned updater. The matched update comparison also improves next-camera accuracy and arrival-time prediction. Its identity-retention advantage grows across repeated handoffs, connecting the better forecast to longer identity continuity. Natural-gap and camera-masking tests examine recovery without fresh observations, while a matched-coverage diagnostic measures the role of waiting in avoiding premature matches.

\Needspace{4\baselineskip}
Our contributions to language-guided multi-camera tracking are as follows.

\noindent\textbf{(1) Learning a predictive state from uncertain associations.} Candidate- and null-conditioned states contribute through posterior weights and a learned recurrent projection. Later handoffs supervise the information retained for future observations across the network.

\noindent\textbf{(2) Coupling prediction, association and waiting.} A shared belief connects event forecasts to identity decisions. Elapsed time and camera availability update this belief during waits, distinguishing an unseen target from an unavailable stream that supplies no observations.

\noindent\textbf{(3) Connecting return prediction to identity continuity.} Matched controls compare structured feedback with fixed moments and a generic learned updater. Improved forecasts accompany stronger identity retention across scenes and repeated handoffs. Missing-observation tests and waiting diagnostics identify the situations in which this continuity is preserved.

\section{Related Work}
\label{sec:related}
\paragraph{Identity memory and uncertain association.}
Graph association links detections over time \citep{neuralgraph,sushi}, while track queries retain identity evidence between frames \citep{trackformer,motr}. Object-permanence models maintain targets beyond the current view \citep{permatrack,osnom}. Whareformer \citep{whareformer} updates appearance and 3D-location memories after assignment. It uses a New Track token for unseen objects. Polycepta \citep{polycepta} learns recurrent object states to predict future appearance. Probabilistic association weights candidate and missed-detection states \citep{pda,jpda}, and multiple-hypothesis tracking retains alternative histories \citep{mht}. \emph{ReWorld-Track learns the predictive state passed between cameras from weighted candidate and null states. Errors in later return predictions supervise this summary.}

\paragraph{Predicting returns across cameras.}
Trajectory Tensors \citep{trajectorytensors} predict the camera, arrival time and location of a future observation. SMO-MCTF \citep{smomctf} extends location-tensor forecasting to multiple objects. Cross-camera feature prediction and camera-conditioned generation anticipate appearance changes using intra-camera supervision \citep{ccfp,camerafeatures}. \emph{ReWorld-Track feeds candidate- and null-conditioned states back into the next camera, time and entry forecast. This update lets the route forecast retain alternative locations and motions from ambiguous returns across successive camera transitions.}

\paragraph{Language as identity evidence.}
Retrieval and referring tracking use descriptions to identify targets \citep{cityflownl,rmot,ikun,ovtr}. Visual and image--text representations support this matching \citep{dino,clip,blip}. CRTracker combines views to recover hidden attributes \citep{crmot}. ViewSAM learns view-aware semantics with weak supervision \citep{viewsam}, and LaMMOn combines language representations with online graph association across cameras \citep{nguyen2024lammon}. A description can still fit several candidates or name an attribute hidden in the current view. \emph{ReWorld-Track uses its graded support alongside appearance and the event prior. Query evidence changes the candidate weights and thus the association state passed to the next event forecast.}

\paragraph{Predictive latent dynamics and world models.}
World models learn latent predictive states that summarize observation history for forecasting future observations and supporting downstream decisions \citep{worldmodels,planet,dreamer}. More recent work further emphasizes closed-loop prediction and state updating as new evidence arrives \citep{worldinworld}. \emph{ReWorld-Track adopts this predictive-state perspective for multi-camera tracking, but models structured target-observation events rather than future pixels. Its state predicts the next camera, arrival time and entry region, and is recursively revised by uncertain association outcomes.}

\section{Recursive Event World Model}
\label{sec:method}
\subsection{From a query to successive handoffs}
A query $q$ and source observation initialize a persistent identity $g$ to follow across camera changes and absence. At time $t$, the tracker receives camera-local candidates $y_j\in\Cand_t$, each with a timestamp, box, appearance and observed track prefix. The history $\Hist_t$ contains only observations received up to $t$. The tracker either links a candidate to the identity or continues waiting. Its belief records the possible target states behind these decisions. Prediction advances this belief through the gap, association evaluates the current candidates, and feedback prepares the state for the next observation.

\begin{figure}[!htb]
\centering
\includegraphics[width=\linewidth]{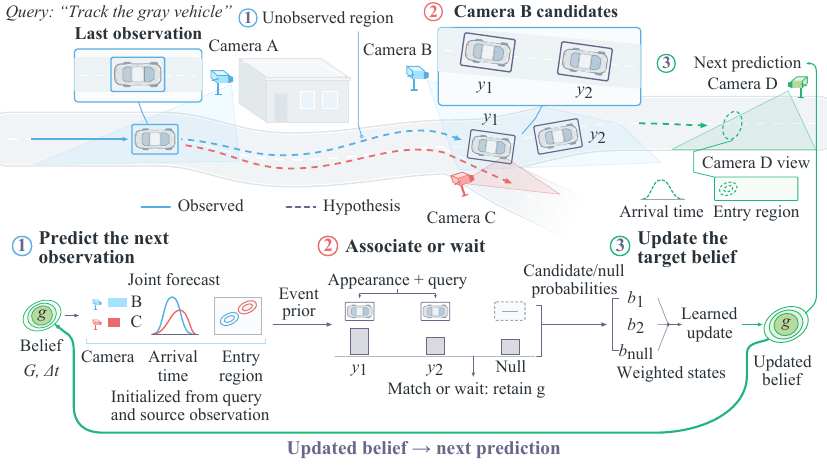}
\caption{Overview of \method{}. The scene connects the three steps to successive camera observations. Prediction guides candidate-or-null association. The resulting posterior updates the target belief. The green loop carries uncertainty into the next forecast. Bars and curves are illustrative.}
\label{fig:method}
\end{figure}

In Figure~\ref{fig:method}, prediction after departure from A supports possible returns at B and C. At B, association combines this forecast with appearance and query evidence. The candidate and null probabilities weight the states for $y_1$, $y_2$ and null to update the belief for the next return at D.

\subsection{Predicting the next observation}
\label{sec:belief}
Prediction assigns probabilities to the possible states after the last observation. Each state is $z_t=(u,\ell_t,v_t,e_t)$. Its components are identity attributes $u$, camera or blind-transition location $\ell_t$, motion $v_t$ and network presence $e_t$. A fixed graph $G=(\Camera,\mathcal E)$ specifies camera transitions. Recursive filtering \citep{kalman,particle} propagates the posterior over elapsed time $\Delta t=t-t^-$,
\begin{equation}
 b_t^-(z)=\int K_\theta(z\given z',G,\Delta t)b_{t^-}^+(z')\,dz',
 \qquad b_t^+(z)=p(z_t=z\given\Hist_t).
 \label{eq:propagate}
\end{equation}
Minus and plus denote prior and posterior beliefs. We represent location with a categorical distribution over cameras and blind-transition regions. Velocity has a mean and diagonal covariance. A 256-dimensional recurrent state summarizes identity and observation history. A 256-unit gated recurrent unit (GRU) and two graph-message layers implement $K_\theta$. They receive these quantities, network presence, installation features and elapsed time. The graph constrains possible routes, while the recurrent state adapts their probabilities to the target's history.

The event decoder uses the updated belief to forecast the next observation \citep{rmtpp,thp}. Each event has a camera $c$, arrival time $\tau>0$ and entry position $r$. We measure $r\in[0,1]^2$ in normalized image coordinates. The joint forecast factors as
\begin{equation}
 p_\theta(c,\tau,r\given b_t^+,G,e=1)
 =p_\theta(c\given b_t^+,G,e=1)\,
 p_\theta(\tau\given c,b_t^+,G,e=1)\,
 p_\theta(r\given c,\tau,b_t^+,G,e=1).
 \label{eq:event}
\end{equation}
Routes toward B and C in Figure~\ref{fig:method} imply different arrival times and entry positions. A categorical distribution models the next camera. Each camera has a three-component log-normal arrival mixture and a two-component Gaussian entry mixture in logit space. These mixtures allow several plausible returns within one camera. Conditioning time and entry position on the destination associates each route with its expected arrival time and image-space entry region.

If the predicted arrival fails to occur, availability $a_c(t)\in\{0,1\}$ controls the evidence provided by continued absence. Failure to detect the target in an available camera can weaken that route. A masked stream supplies no observation evidence, though the target can traverse its region. The probability $s_t=p(e_t=1\given\Hist_t)$ tracks whether the target remains in the network during unobserved travel. In both cases, the physical state continues to evolve even without a new image of the target.

\subsection{Associating candidates or continuing to wait}
\label{sec:association}
At each 10\,Hz update, the preceding forecast assigns a prior mass $\eta_j$ to each candidate. We integrate the event density over the candidate's camera, time and disjoint entry region. Overlapping supports are allocated to the nearest entry centroid to avoid counting the same event mass twice. Stream availability and learned detection opportunity determine the observable share of this mass. For an active target, the remaining mass $\eta_0=1-\sum_j\eta_j$ represents no match among the current candidates. It includes unavailable views, missed detections and returns outside the candidate supports.

The event prior narrows the plausible returns, but candidates at similar places and times still require appearance and language evidence. Frozen Contrastive Language--Image Pretraining (CLIP) features \citep{clip} pass through trainable 256-dimensional projections. Appearance averages the latest eight causal crop features, and the source-history query is encoded once. A logistic scorer combines appearance and query similarities with motion residual and detector confidence. Its validation-calibrated logit gives the likelihood ratio $L_j$ after exponentiation. Normalizing with the priors gives posterior probabilities over the candidates and the temporary-null alternative
\begin{equation}
 W_t=\eta_0+\sum_{k=1}^{|\Cand_t|}\eta_kL_k,\qquad
 \beta_j=\frac{\eta_jL_j}{W_t},\qquad
 \beta_0=\frac{\eta_0}{W_t}.
 \label{eq:association}
\end{equation}
Null has likelihood ratio one and is conditioned on an active target. It represents no match among current candidates. Network presence separately describes whether the target remains in the camera network. An implausible arrival weakens an appearance match, while weak candidates leave support on null. The bars in Figure~\ref{fig:method} depict $\beta_1$, $\beta_2$ and $\beta_0$.

Acceptance requires a posterior threshold and margin over the next candidate for two consecutive updates. Per-camera maximum-weight matching resolves competing identities. An empty set yields null. Camera-local association lets one view wait while another observes the identity. Same-time posteriors share one prior and combine with normalized detector-confidence weights.

\subsection{Carrying uncertainty into the next forecast}
\label{sec:feedback}
The match-or-wait decision gives the current tracking output. Feedback prepares the next forecast by retaining evidence from all plausible alternatives. Conditioning on candidate $y_j$ gives a state $b_{t,j}$ with its location, motion and identity evidence. Conditioning on continued absence gives the null state $b_{t,0}$. Probabilistic association \citep{pda,jpda} weights these alternatives,
\begin{equation}
 b_t^+(z\given e_t=1)=\beta_0b_{t,0}(z)+\sum_j\beta_jb_{t,j}(z).
 \label{eq:update}
\end{equation}
At B in Figure~\ref{fig:method}, these components are $b_1$, $b_2$ and $b_{\mathrm{null}}$. Plausible candidates contribute according to their posterior weights. During a wait, the null component carries the active identity forward. Its state accounts for continued absence and time since the last observation.

We summarize the mixture in a fixed-size state. Posterior-weighted region probabilities retain alternative routes, and velocity moments describe motion uncertainty. A learned projection summarizes the weighted conditioned states in the recurrent representation. The GRU propagates these quantities and network presence for the event decoder. Commitment reads this posterior but preserves the weighted state update. Candidate and null evidence therefore shape the next forecast during waits.

Keeping these statistics explicit gives the predictor direct access to route and motion uncertainty. The learned projection adds a trainable summary of conditioned identity states. Fixed-moment feedback omits this learned summary. A generic set encoder maps the same posterior information directly into a recurrent state. ReWorld retains the weighted hypothesis structure alongside that state. Table~\ref{tab:core}(b) compares these updates under the same predictor, scorer, training data and rollout depth.

Prediction and feedback follow the observation order. The preceding posterior forecasts the current return. Current candidate likelihoods then revise the state used for the following return. At B in Figure~\ref{fig:method}, uncertainty between the two vehicles changes the forecast at D. Later evidence can resolve an earlier ambiguity through this retained predictive state.

\subsection{Learning from successive returns}
Training connects the state update to its purpose: predicting later observations and preserving identity. Camera and association losses use cross-entropy, including null as an association outcome. Arrival time and entry position use negative log-likelihood. A window ending without an observation contributes a right-censored waiting likelihood \citep{kaplanmeier}. This term trains the probability that the target remains unseen throughout the observed interval. Binary presence and contrastive identity supervision further constrain the shared state.

Each training window contains 32 observation updates and at most four handoffs. Gradients cross those handoffs, allowing a later forecast error to train an earlier posterior projection. Teacher forcing decreases over the first 20 epochs, exposing the update to its own uncertain associations. Gradients are detached between windows, and inference uses fixed trained parameters. Appendix~\ref{app:method-case} follows a complete vehicle update. Appendix~\ref{app:implementation} gives the full objective and training schedule.

\FloatBarrier
\section{Experiments}
\label{sec:experiments}
We test whether the predictive state preserves identity over complete trajectories, then examine the update mechanism and the difficult returns it must handle.

\subsection{Experimental setup}
CityFlowV2 vehicles \citep{cityflow,cityflowv2} and MTMMC pedestrians \citep{mtmmc} use separate RGB training and scene-disjoint tests. For MTMMC, we add source-history language descriptions using the annotation protocol in Appendix~\ref{app:interventions}. Query construction uses only observations available at initialization and excludes all future target observations. Table~\ref{tab:main} gives the fixed evaluation counts. Ground-truth visibility is used only to define evaluation eligibility and deadlines. Detector misses remain eligible, and an incorrect first commitment counts as failure. Appendix~\ref{app:protocol-records} defines the handoff events and absence decisions used in this evaluation.

\begin{table}[!b]
\centering
\caption{Identity continuity with separate training by domain. Rates are percentages, and bold marks column optima. ReID denotes re-identification, BoT person-ReID and HC hierarchical clustering. $\dagger$ marks our causal adaptation of an external tracker. Its results are re-evaluated with the causal initialization, test events and trajectory metrics specified in Section~\ref{sec:experiments}.}
\label{tab:main}
\small
\setlength{\tabcolsep}{3pt}
\begin{tabularx}{\linewidth}{@{}l*{7}{>{\raggedleft\arraybackslash}X}@{}}
\toprule
\rowcolor{rwHeader}
\bfseries Method & \bfseries HA $\uparrow$ & \bfseries IDF1 $\uparrow$ & \bfseries HOTA $\uparrow$ & \bfseries DetA $\uparrow$ & \bfseries AssA $\uparrow$ & \bfseries IDSW $\downarrow$ & \bfseries FM $\downarrow$ \\
\midrule
\rowcolor{rwGroupBlue}
\multicolumn{8}{@{}l}{\textit{CityFlowV2}: $D_+=3587$, $D_0=14291$} \\
ReID & 71.81 & 68.54 & 51.86 & 57.46 & 46.81 & 326 & 9.99 \\
ReID + time & 77.03 & 72.89 & 56.72 & 57.53 & 55.91 & 249 & 5.86 \\
Reactive memory & 80.23 & 76.62 & 61.48 & 57.68 & 65.53 & 191 & 6.72 \\
TrackTA$^{\dagger}$ & 79.15 & 75.48 & 60.63 & 57.31 & 64.15 & 237 & 5.28 \\
LaMMOn$^{\dagger}$ & 82.27 & 79.06 & 64.31 & \textbf{57.82} & 71.53 & 174 & 4.95 \\
Camera-link & 82.91 & 78.69 & 63.94 & 57.69 & 70.87 & 182 & 4.61 \\
\rowcolor{rwMethod}
\textbf{ReWorld-Track} & \textbf{84.67} & \textbf{80.37} & \textbf{65.19} & 57.76 & \textbf{73.58} & \textbf{158} & \textbf{4.28} \\
\midrule
\rowcolor{rwGroupPurple}
\multicolumn{8}{@{}l}{\textit{MTMMC}: $D_+=2749$, $D_0=10683$} \\
ReID & 66.42 & 41.76 & 29.83 & 43.62 & 20.40 & 589 & 11.17 \\
ReID + time & 71.26 & 46.82 & 34.29 & 43.78 & 26.86 & 463 & 6.39 \\
Reactive memory & 77.41 & 52.17 & 40.16 & 44.03 & 36.63 & 391 & 7.74 \\
TrackTA$^{\dagger}$ & 74.61 & 49.86 & 37.41 & 43.71 & 32.02 & 427 & 6.14 \\
QDTrack + BoT + HC$^{\dagger}$ & 78.06 & 53.74 & 41.02 & 44.16 & 38.10 & 368 & 6.82 \\
CRTracker$^{\dagger}$ & 81.59 & 55.91 & 44.12 & 44.31 & 43.92 & 339 & 5.67 \\
\rowcolor{rwMethod}
\textbf{ReWorld-Track} & \textbf{83.56} & \textbf{58.12} & \textbf{45.36} & \textbf{44.37} & \textbf{46.39} & \textbf{298} & \textbf{4.90} \\
\bottomrule\end{tabularx}\end{table}

Identity F1 score (IDF1) and higher order tracking accuracy (HOTA) assess complete trajectories, including cross-camera identity continuity \citep{idmetrics,hota}.

\begin{samepage}
Detection accuracy (DetA) and association accuracy (AssA) separate detection from association quality. Identity switches (IDSW) further locate identity-continuity errors.\par
\end{samepage}

Handoff accuracy (HA) measures correct returns, while false-match rate (FM) measures distractor acceptance during absence. Delay measures time to correct acceptance. Internal controls share detections, causal prefixes and frozen CLIP features. TrackTA follows MTMMC \citep{mtmmc}. QDTrack + BoT + HC combines QDTrack \citep{qdtrack}, Bag-of-Tricks ReID \citep{bagtricks} and hierarchical clustering. Five paired training seeds use validation-selected checkpoints and thresholds. Appendix~\ref{app:protocol-records} details the causal adaptations and training settings.

\subsection{Identity continuity in vehicle and pedestrian domains}
\label{sec:principal}
The same full ReWorld-Track configuration is evaluated separately on the vehicle and pedestrian domains (Table~\ref{tab:main}). On vehicles, it recovers 159 more handoffs than reactive memory with 349 fewer false accepts. AssA rises from 65.53\% to 73.58\%, while detection accuracy remains similar. Fewer false commitments protect identity while additional correct returns extend its trajectory.

Pedestrians show the same association benefit. Against adapted CRTracker, AssA rises from 43.92\% to 46.39\%, with DetA almost unchanged at 44.31\% and 44.37\%. Across five paired training seeds, mean HOTA improves by 1.24 points on the fixed test set. The 95\% Student-$t$ interval over seed-wise differences is [1.10, 1.38]. The gains in both domains concern association quality.

Camera-link, an internal fixed-prior baseline, uses training-derived transition frequencies and travel times, reaching 63.94 vehicle HOTA versus 65.19 for ReWorld-Track. We therefore examine how recent evidence revises each target's next-camera and arrival forecast.

\begin{figure}[!t]
\centering
\includegraphics[width=.86\linewidth,trim=0 1.5bp 0 0,clip]{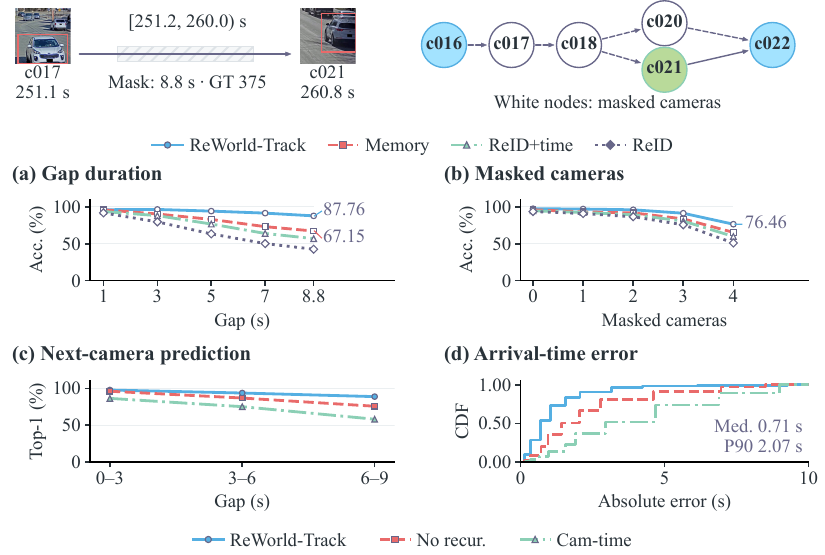}
\caption{Missing-observation experiments. The GT375 timeline and six-camera graph illustrate masking. (a,b) Handoff accuracy versus mask duration and camera count. (c) Next-camera prediction by gap. (d) Arrival-time error CDF. No recur. omits feedback. Cam-time predicts camera and time.}
\label{fig:sensitivity}
\medskip
\includegraphics[width=.86\linewidth,trim=0 1.5bp 0 0,clip]{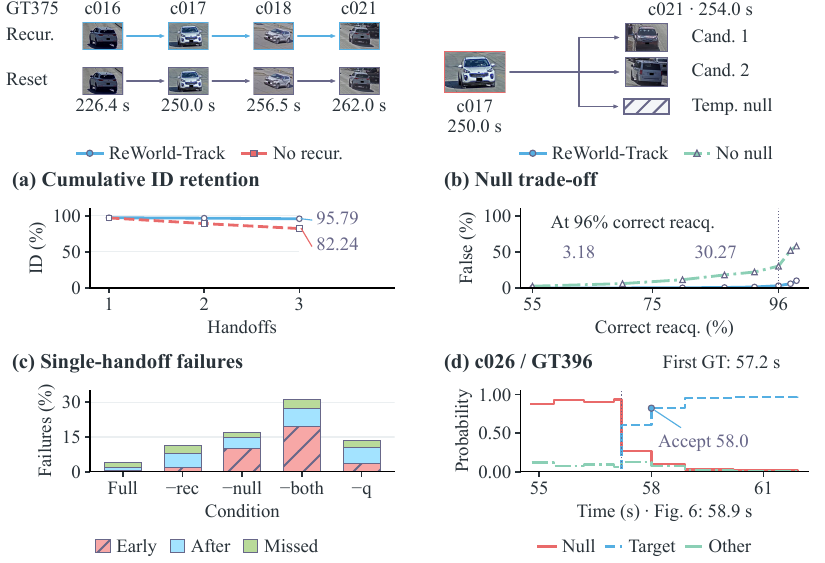}
\caption{Feedback and waiting. Upper examples show GT375 across repeated handoffs and a choice between candidates and temporary null. (a) Identity retention. (b) Diagnostic false-match comparison at matched test-set reacquisition. (c) Handoff failures. (d) The c026-local GT396 belief from appearance at 57.2\,s to acceptance at 58.0\,s, with rising target support.}
\label{fig:mechanisms}
\end{figure}

\subsection{How feedback preserves identity}
\label{sec:mechanism-results}
The prior and feedback address different errors. Removing the prior lowers MTMMC HOTA from 45.36 to 40.02 (Table~\ref{tab:core}(a)), showing the value of destination and timing information during association. Restarting each forecast from the last committed observation gives 41.28. This control retains appearance memory and prediction heads but discards intervening posterior updates. The drop shows the value of retaining gap observations before a reliable match becomes available.

\begin{table}[!htbp]
\centering
\caption{MTMMC controls on 2749 handoffs and 10683 absence decisions. (a) Means and sample standard deviations over five seeds. (b) Three posterior updates under the same event predictor. Rates are percentages. Delay is in seconds. \mbox{w/o means without.}}
\label{tab:core}
\small\setlength{\tabcolsep}{3pt}
\textbf{(a) Component ablations}\par
\begin{tabularx}{\linewidth}{@{}>{\raggedright\arraybackslash}Xrrrrr@{}}
\toprule
\rowcolor{rwGroupBlue}
\bfseries Variant & \bfseries HA $\uparrow$ & \bfseries IDF1 $\uparrow$ & \bfseries HOTA $\uparrow$ & \bfseries FM $\downarrow$ & \bfseries Delay (s) $\downarrow$ \\
\midrule
\rowcolor{rwMethod}
\textbf{ReWorld-Track} & 83.56 $\pm$ 0.23 & 58.12 $\pm$ 0.24 & 45.36 $\pm$ 0.20 & 4.90 $\pm$ 0.15 & 1.03 $\pm$ 0.02 \\
w/o event prior & 77.20 $\pm$ 0.23 & 52.61 $\pm$ 0.24 & 40.02 $\pm$ 0.19 & 7.41 $\pm$ 0.19 & 1.39 $\pm$ 0.03 \\
w/o recursion & 78.74 $\pm$ 0.25 & 54.36 $\pm$ 0.23 & 41.28 $\pm$ 0.19 & 6.81 $\pm$ 0.19 & 1.26 $\pm$ 0.03 \\
w/o temporary null & 85.05 $\pm$ 0.22 & 57.54 $\pm$ 0.20 & 44.98 $\pm$ 0.21 & 12.74 $\pm$ 0.25 & 0.61 $\pm$ 0.02 \\
w/o query & 81.72 $\pm$ 0.24 & 56.79 $\pm$ 0.21 & 44.10 $\pm$ 0.19 & 5.42 $\pm$ 0.17 & 1.09 $\pm$ 0.02 \\
\bottomrule\end{tabularx}
\end{table}

\noindent\begin{minipage}[t]{.425\linewidth}
\vspace{0pt}
Table~\ref{tab:core}(b) tests what the update should retain. A generic learned updater improves on fixed moments. ReWorld-Track gains a further 0.50 HOTA points with fewer false matches. The two learned updates receive the same posterior information and use 0.64M and 0.66M parameters (Table~\ref{tab:update-cost}).
\end{minipage}\hfill
\begin{minipage}[t]{.545\linewidth}
\vspace{0pt}
\small\centering\textbf{Table~\ref{tab:core}(b): Posterior update mechanisms}\par\smallskip
\setlength{\tabcolsep}{2.5pt}
\begin{tabularx}{\linewidth}{@{}>{\raggedright\arraybackslash}Xrrrr@{}}
\toprule
\rowcolor{rwGroupPurple}
\bfseries Update & \bfseries HA $\uparrow$ & \bfseries IDF1 $\uparrow$ & \bfseries HOTA $\uparrow$ & \bfseries FM $\downarrow$ \\
\midrule
\shortstack[l]{Fixed-moment\\soft association} & 82.25 & 57.06 & 44.42 & 5.53 \\
\shortstack[l]{Generic learned\\updater} & 82.89 & 57.54 & 44.86 & 5.19 \\
\rowcolor{rwMethod}
\shortstack[l]{\bfseries ReWorld\\\bfseries posterior} & \textbf{83.56} & \textbf{58.12} & \textbf{45.36} & \textbf{4.90} \\
\bottomrule\end{tabularx}

\end{minipage}

The prediction results help explain this tracking gain. Relative to the generic updater, next-camera accuracy rises from 86.03\% to 87.41\%, and median arrival error falls from 0.78 to 0.71\,s (Table~\ref{tab:update-forecast}). Better forecasts narrow the plausible returns at the next association. The HOTA advantage also persists across three independent scenes at 0.46--0.50 points (Table~\ref{tab:update-scenes}).

Waiting matters because a locally faster decision can damage the complete trajectory. Removing temporary null raises HA to 85.05\% and reduces mean delay from 1.03 to 0.61\,s. Yet false matches rise from 4.90\% to 12.74\%, and both IDF1 and HOTA decline. Earlier acceptance admits more distractors during absence. Null lets evidence accumulate before commitment, trading a short wait for fewer identity errors. Tables~\ref{tab:results5} and~\ref{tab:supp26} examine calibration and the recovery--delay tradeoff.

\subsection{Missing observations and repeated handoffs}
\label{sec:missing-results}
Figure~\ref{fig:sensitivity} tests the belief between sightings by varying mask duration and camera count in a six-camera vehicle network. At an 8.8\,s mask, recovery reaches 87.76\% versus 67.15\% for reactive memory. The forecast panels show improved next-camera accuracy across gap bins and a median arrival error of 0.71\,s. The belief preserves a useful forecast as the last visual observation becomes older.

Natural gaps test the same need during ordinary travel between views, with all streams available. Masking additionally removes selected streams, detections and features. Both favor feedback, and the advantage over reactive memory grows for longer natural gaps (Tables~\ref{tab:results3} and~\ref{tab:supp27}).

The difference widens over repeated handoffs. Identity retention reaches 95.79\% with feedback versus 82.24\% with restarted prediction in Figure~\ref{fig:mechanisms}(a). Against the generic learned updater, ReWorld-Track's advantage grows from 0.27 points after one handoff to 2.65 after three (Table~\ref{tab:update-retention}). This widening gap shows that update quality increasingly affects identity retention across successive handoffs.

The waiting diagnostic holds recovery coverage fixed. At 96\% attained reacquisition in the test-set sweep, false matches are 3.18\% with null and 30.27\% without it (Figure~\ref{fig:mechanisms}(b)). Deployment uses validation-selected thresholds. Null avoids premature distractor commitments at equal recovery coverage. Panel~(d) follows GT396 through this decision. During the 0.8\,s between visibility at 57.2\,s and acceptance at 58.0\,s, rising target support resolves the initially uncertain return.

\FloatBarrier

\subsection{Language evidence and generalization}
\label{sec:language-transfer}
\noindent\begin{minipage}[t]{.425\linewidth}
\vspace{0pt}
Language is most useful when appearance leaves a close competitor. Table~\ref{tab:language} tests controlled changes to source-history descriptions on fixed identities and observations. Visibility annotations define the partly unobservable condition. NQ omits the query, HG uses a hard attribute gate, and F uses the full scorer. The gate rejects confidently observed contradictions and treats unknown attributes as neutral. The full scorer combines graded query similarity with appearance and the event prior.
\end{minipage}\hfill
\begin{minipage}[t]{.545\linewidth}
\vspace{0pt}
\begin{rwtableside}
\caption{Controlled query perturbations on 637 vehicle and 509 pedestrian events. Values are HA (\%). Bold marks each row's best result for that condition.}
\label{tab:language}
\small\setlength{\tabcolsep}{2.6pt}
\begin{tabularx}{\linewidth}{@{}l*{3}{>{\raggedleft\arraybackslash}X}@{}}
\toprule
\rowcolor{rwHeader}
\bfseries Query condition & \bfseries NQ & \bfseries HG & \bfseries F \\
\midrule
\rowcolor{rwGroupBlue}
\multicolumn{4}{@{}l}{\textit{CityFlowV2}} \\
Natural description & 81.63 & 85.40 & \textbf{87.76} \\
Partly unobservable & 81.63 & 74.10 & \textbf{84.46} \\
Misleading attributes & \textbf{81.63} & 65.31 & 75.67 \\
\rowcolor{rwGroupPurple}
\multicolumn{4}{@{}l}{\textit{MTMMC}} \\
Natural description & 78.39 & 80.16 & \textbf{84.28} \\
Partly unobservable & 78.39 & 68.96 & \textbf{81.14} \\
Misleading attributes & \textbf{78.39} & 64.05 & 73.08 \\
\bottomrule\end{tabularx}\end{rwtableside}

\end{minipage}

Natural descriptions improve recovery in both domains. With partly unobservable attributes, the full scorer remains above NQ while the hard gate falls below it. Graded evidence can still support a match when part of a description is hidden by the current view. On visually ambiguous MTMMC candidates, natural queries add 10.9 HA points, compared with 2.1 on easy cases (Table~\ref{tab:results4}). Language complements the forecast when several candidates fit the predicted return. Misleading descriptions harm both query-based methods by favoring an incorrect identity.

We next test whether the learned predictive update remains useful under object-domain and installation shifts. The transferred model uses frozen public encoders and the target installation graph in both directions. With source-trained checkpoints and thresholds, ReWorld-Track improves HA over appearance-plus-time association by 4.61 points from vehicles to pedestrians and 5.22 in reverse. False matches also decrease (Table~\ref{tab:cross-domain}, Appendix~\ref{app:transfer-resources}). On unseen MTMMC topology, HA falls from 85.1\% to 74.6\% but remains 3.8 points above one-way prediction (Table~\ref{tab:results7}). Both methods receive the target installation graph. Feedback remains useful on unfamiliar routes.

\section{Conclusion}
ReWorld-Track learns a \textbf{predictive state from uncertain identity assignments}. Posterior weights and a learned recurrent projection carry candidate and absence evidence into later forecasts. Matched controls connect this update to \textbf{stronger identity continuity across independent scenes and repeated handoffs}. On a supplied camera graph, the belief links recovery after absence to the history needed for later returns. Misleading descriptions and unfamiliar route statistics reduce recovery, leaving accurate identity continuation under these shifts as an open challenge.
\label{maintextend}

\FloatBarrier
\clearpage
\section*{AI Use Statement}
Generative AI tools were used to assist with literature research, experimental support, and language polishing. All AI-assisted content was reviewed and verified by the authors, who take full responsibility for the final manuscript and reported results.

\section*{Ethics Statement}
Multi-camera tracking links observations across locations and can increase surveillance capabilities. Deployment requires appropriate authorization, access controls, and limits on data retention and downstream use. This study uses vehicle and pedestrian benchmark imagery under the corresponding access terms. Benchmark identities are used for evaluation, without assigning real-world names. Language and visual evidence can be incomplete or biased. Uncertain predictions require care in consequential applications, where a wrong association could affect an individual.

\section*{Reproducibility Statement}
Section~\ref{sec:method} defines the recursive model, and Appendix~\ref{app:implementation} provides its architecture, learning schedule and concurrent-view update. Appendices~\ref{app:protocol-records}--\ref{app:additional-controls} and~\ref{app:update-controls} document the evaluation populations, baseline adaptations, annotations, transfer resources and update comparisons. Each protocol identifies the events and sampling units used in its comparisons. The source package includes eight figure PDFs, numerical data and deterministic table builders for the reported comparisons.

\bibliography{references}
\bibliographystyle{iclr2027_conference}
\clearpage
\appendix
\section{Pedestrian Identity Continuation}
\label{app:pedestrian-cases}
\begin{figure}[!htb]
\centering
\includegraphics[width=.88\linewidth,trim=0 12bp 0 1bp,clip]{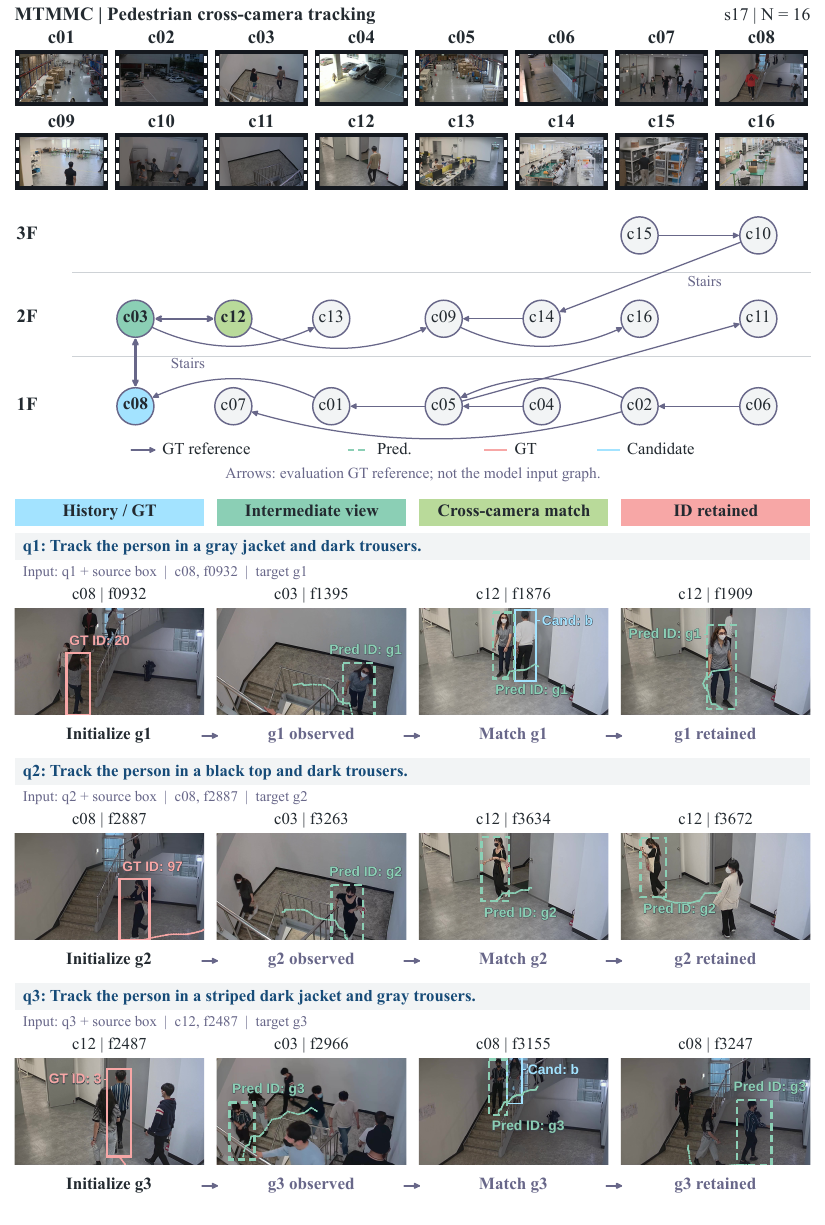}
\caption{Three query-initialized MTMMC s17 tracks with all sixteen streams available. The rows show GT20, GT97 and GT3 as $g_1$, $g_2$ and $g_3$. Columns show initialization, an intermediate observation, matching and continuation. The floor graph marks annotated transitions. Rose boxes show GT, mint dashed boxes predictions and blue boxes alternatives.}
\label{fig:pedestrians}
\end{figure}

Figure~\ref{fig:pedestrians} follows three query-initialized identities on MTMMC s17. GT20 and GT97 move through c08--c03--c12. GT3 takes the reverse route. Each return changes the view of the query attributes and the evidence for matching the same identity across the camera network.

\FloatBarrier
\clearpage

\clearpage
\section{A Vehicle Example of the Recursive Update}
\label{app:method-case}
\begin{figure}[!htb]
\centering
\includegraphics[width=\linewidth]{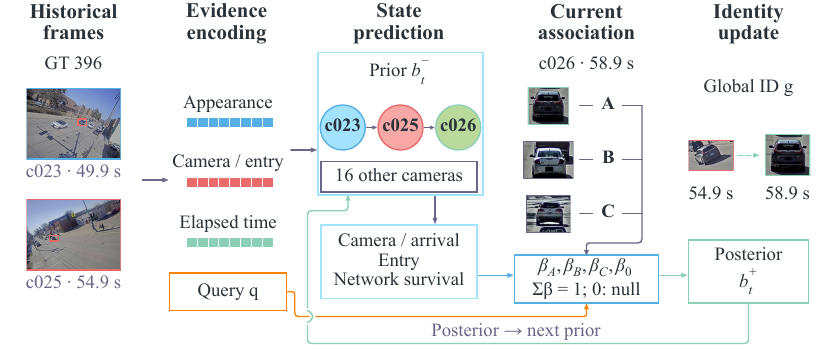}
\caption{A vehicle instance of prediction, association and posterior feedback. The c023 and c025 observations condition the forecast, and the c026 candidates at 58.9\,s supply new evidence. The posterior updates the persistent target state for the next handoff. Camera labels and times refer to this sequence. Figure~\ref{fig:method} presents the corresponding general framework.}
\label{fig:method-case}
\end{figure}

Figure~\ref{fig:method-case} instantiates the three steps of Figure~\ref{fig:method} for GT396 in the nineteen-camera network. The observations in c023 at 49.9\,s and c025 at 54.9\,s establish the appearance, location and elapsed-time history. The query identifies the gray vehicle whose persistent identity is $g$.

The \emph{State prediction} column forecasts the next camera, arrival time and entry region from this history, together with continued network presence. At 58.9\,s, the \emph{Current association} column compares candidates A, B and C in c026 with the event forecast and query. Candidate A continues $g$. In the \emph{Identity update} column, candidate and null probabilities weight their conditioned states. The green loop carries this evidence into the next prediction. Keeping these alternatives in the update lets later observations refine an identity association that remains uncertain at this return.

The 58.9\,s image shows the accepted continuation. The earlier decision trace in Figure~\ref{fig:mechanisms}(d) records first appearance at 57.2\,s and acceptance at 58.0\,s. These views connect the state update to the observations used in the vehicle analysis, with the timestamp conventions detailed in Appendix~\ref{app:metrics}.

\section{Language-Conditioned Vehicle Cases}
\label{app:vehicle-cases}
\begin{figure}[!htbp]
\centering
\includegraphics[width=.80\linewidth]{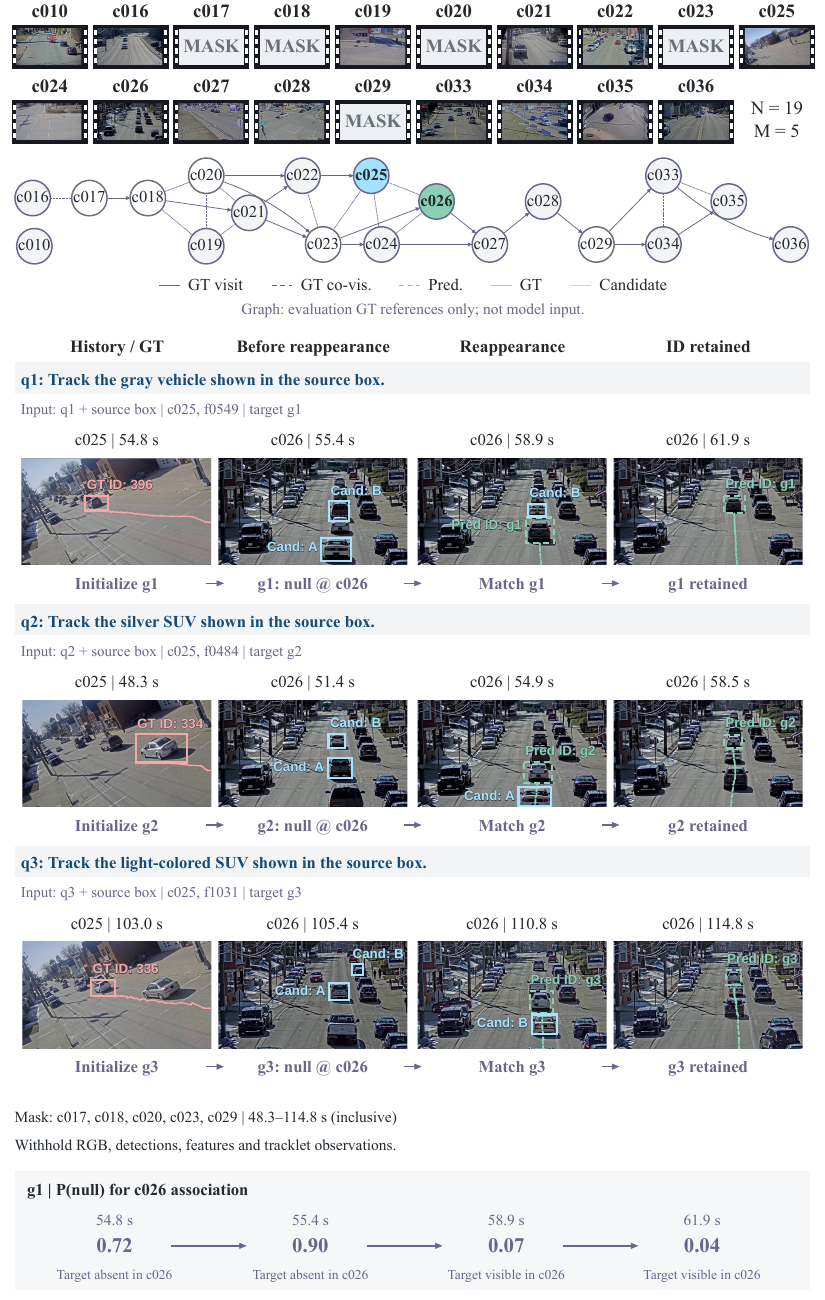}
\caption{Language-conditioned vehicle association in CityFlowV2 S05. A query and source box initialize each persistent identity. The columns follow c026 before reappearance, after matching and during continuation. Rose GT ID labels identify targets 396/334/336, mint Pred ID labels show $g_1/g_2/g_3$, and blue candidate letters identify alternatives within each image. Dots sample historical trails. Of $N=19$ cameras, $M=5$ are masked over nominal time 48.3--114.8\,s inclusive, withholding RGB, detections, features and tracklet observations. The graph displays annotated visit order and co-visibility, highlighting the available c025/c026 views. For $g_1$, the early null probabilities concern c026 while the target remains annotated and visible in the available c025 stream.}
\label{fig:qualitative}
\end{figure}

Figure~\ref{fig:qualitative} follows waiting and matching for query-initialized vehicles GT396, GT334 and GT336. For $g_1$, c026-local $P(\mathrm{null})$ is 0.72, 0.90, 0.07 and 0.04 at 54.8, 55.4, 58.9 and 61.9\,s. Early snapshots favor null in c026 while GT396 is still visible in available c025. Support shifts to the c026 candidate once the vehicle appears there. The graph places these views in the network. Appendix~\ref{app:enlarged} enlarges all twelve panels for inspection of candidate boxes and identity continuation.

\FloatBarrier
\clearpage

\clearpage
\section{Enlarged Language-Conditioned Vehicle Cases}
\label{app:enlarged}
The enlarged panels from Figure~\ref{fig:qualitative} follow each query-initialized identity from its source view through waiting, matching and continuation. Rose GT ID labels identify annotated targets, mint Pred ID labels show persistent display identities, and blue candidate letters identify image-local alternatives. Dots show historical observations available at each frame. The null stage concerns c026 association, so a target can remain visible elsewhere in the network.
\subsection{GT396 with display identity \texorpdfstring{$g_1$}{g1}}
The initialization query is ``Track the gray vehicle shown in the source box.''

\begin{minipage}[t]{0.48\linewidth}
\centering
Source-box initialization\\
c025, 54.8\,s\par\smallskip
\includegraphics[width=\linewidth,trim={10.725000 352.953125 294.395000 219.846875},clip]{fig5_v19.pdf}
\end{minipage}\hfill
\begin{minipage}[t]{0.48\linewidth}
\centering
Before reappearance\\
c026, 55.4\,s\par\smallskip
\includegraphics[width=\linewidth,trim={104.775000 352.953125 200.345000 219.846875},clip]{fig5_v19.pdf}
\end{minipage}

\par\medskip
\begin{minipage}[t]{0.48\linewidth}
\centering
Cross-camera match\\
c026, 58.9\,s\par\smallskip
\includegraphics[width=\linewidth,trim={200.475000 352.953125 104.645000 219.846875},clip]{fig5_v19.pdf}
\end{minipage}\hfill
\begin{minipage}[t]{0.48\linewidth}
\centering
Identity continuation\\
c026, 61.9\,s\par\smallskip
\includegraphics[width=\linewidth,trim={294.525000 352.953125 10.595000 219.846875},clip]{fig5_v19.pdf}
\end{minipage}

\par\medskip
The source frame initializes $g_1$ for the gray vehicle. At 55.4\,s, the tracker rejects the current c026 alternatives while GT396 remains annotated in available c025. The match shown at 58.9\,s is followed by continuation at 61.9\,s. The diagnostic below follows the change from waiting to matching in c026, alongside the continuing observation in c025. Figure~\ref{fig:mechanisms}(d) supplies the first-appearance and acceptance timing, and Table~\ref{tab:results5} measures aggregate calibration.

\begin{center}
\includegraphics[width=\linewidth,trim={0 0 0 562},clip]{fig5_v19.pdf}
\end{center}

\Needspace{180pt}
\subsection{GT334 with display identity \texorpdfstring{$g_2$}{g2}}
The initialization query is ``Track the silver SUV shown in the source box.''

\begin{minipage}[t]{0.48\linewidth}
\centering
Source-box initialization\\
c025, 48.3\,s\par\smallskip
\includegraphics[width=\linewidth,trim={10.725000 236.953125 294.395000 335.846875},clip]{fig5_v19.pdf}
\end{minipage}\hfill
\begin{minipage}[t]{0.48\linewidth}
\centering
Before reappearance\\
c026, 51.4\,s\par\smallskip
\includegraphics[width=\linewidth,trim={104.775000 236.953125 200.345000 335.846875},clip]{fig5_v19.pdf}
\end{minipage}

\par\medskip
\begin{minipage}[t]{0.48\linewidth}
\centering
Cross-camera match\\
c026, 54.9\,s\par\smallskip
\includegraphics[width=\linewidth,trim={200.475000 236.953125 104.645000 335.846875},clip]{fig5_v19.pdf}
\end{minipage}\hfill
\begin{minipage}[t]{0.48\linewidth}
\centering
Identity continuation\\
c026, 58.5\,s\par\smallskip
\includegraphics[width=\linewidth,trim={294.525000 236.953125 10.595000 335.846875},clip]{fig5_v19.pdf}
\end{minipage}

\par\medskip
The source query and box initialize $g_2$ for GT334 at 48.3\,s. The c026 alternatives at 51.4\,s precede the accepted match shown at 54.9\,s. The final frame follows the same identity to 58.5\,s. Candidate letters identify local alternatives. The silver SUV retains $g_2$ across all four observations.

\Needspace{180pt}
\subsection{GT336 with display identity \texorpdfstring{$g_3$}{g3}}
The initialization query is ``Track the light-colored SUV shown in the source box.''

\begin{minipage}[t]{0.48\linewidth}
\centering
Source-box initialization\\
c025, 103.0\,s\par\smallskip
\includegraphics[width=\linewidth,trim={10.725000 120.953125 294.395000 451.846875},clip]{fig5_v19.pdf}
\end{minipage}\hfill
\begin{minipage}[t]{0.48\linewidth}
\centering
Before reappearance\\
c026, 105.4\,s\par\smallskip
\includegraphics[width=\linewidth,trim={104.775000 120.953125 200.345000 451.846875},clip]{fig5_v19.pdf}
\end{minipage}

\par\medskip
\begin{minipage}[t]{0.48\linewidth}
\centering
Cross-camera match\\
c026, 110.8\,s\par\smallskip
\includegraphics[width=\linewidth,trim={200.475000 120.953125 104.645000 451.846875},clip]{fig5_v19.pdf}
\end{minipage}\hfill
\begin{minipage}[t]{0.48\linewidth}
\centering
Identity continuation\\
c026, 114.8\,s\par\smallskip
\includegraphics[width=\linewidth,trim={294.525000 120.953125 10.595000 451.846875},clip]{fig5_v19.pdf}
\end{minipage}

\par\medskip
For GT336, the source query initializes $g_3$ at 103.0\,s. Local alternatives are visible in c026 at 105.4\,s, before the accepted match shown at 110.8\,s. The 114.8\,s frame then shows continued tracking of the light-colored SUV. As in the other rows, image-local candidate letters change while the display identity remains $g_3$ across the accepted match and the subsequent continuation frame.

\section{Paired Pedestrian Recovery with Missing Cameras}
\label{app:pedestrian-mask}
\begin{figure}[p]
\centering
\includegraphics[width=.78\linewidth]{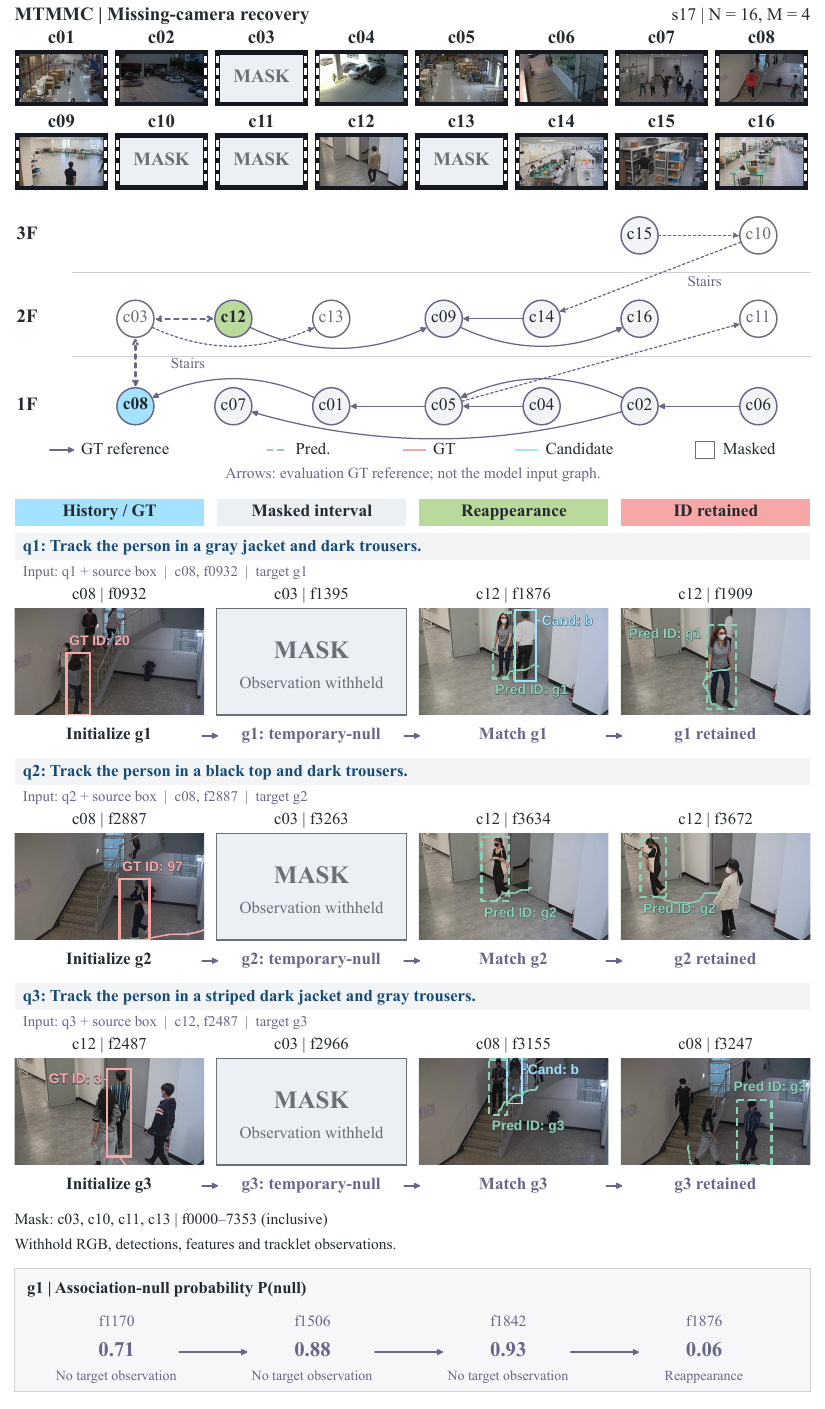}
\caption{Paired pedestrian recovery under missing-camera observations on MTMMC s17. Identities, queries and source boxes match Figure~\ref{fig:pedestrians}. Cameras c03, c10, c11 and c13 are masked over source frames 0--7353 inclusive. This withholds RGB, detections, appearance features and tracklet observations. Prior global state and physical routes remain. Of $N=16$ cameras, $M=4$ are masked, as indicated by the grey panels. The $g_1$ strip follows $P(\mathrm{null})$ among current candidates and temporary null, conditional on an active target, through the gap and reappearance. GT ID and Pred ID identify dataset targets and persistent display identities. Dots sample historical trails. Annotated transitions in the graph provide a reference for tracking under the installation graph.}
\label{fig:pedestrian-mask}
\end{figure}

Figure~\ref{fig:pedestrian-mask} revisits the MTMMC s17 identities, queries and source boxes from Figure~\ref{fig:pedestrians} after intermediate observations are withheld. Four of the sixteen cameras are masked, including c03 in all three cases. Cameras c08 and c12 remain available for initialization and later matching. For $g_1$, association-null probability rises through 0.71, 0.88 and 0.93 at source frames 1170, 1506 and 1842, then falls to 0.06 at frame 1876. The state retains the same identity throughout this transition from waiting to matching. Each row shows the observations that support its eventual continuation.

\section{Observation-Missing Sensitivity and Event Prediction}
\label{app:sensitivity}
Figure~\ref{fig:sensitivity} evaluates handoffs while observations from a six-camera CityFlowV2 network are withheld. At an 8.8\,s mask, HA is 87.76\% for the full model and 67.15\% for reactive memory. Masking four cameras gives full-model HA of 76.46\%. Panels (c,d) examine the forecasts used to recognize the returning target. Arrival-time absolute error has median 0.71\,s and 90th percentile 2.07\,s. Appendix~\ref{app:fixed-mask} extends the masking analysis to the same 600 events at every duration. Table~\ref{tab:results3} tests natural blind gaps, where cameras remain available but the target is out of view.

The preceding masking results use the six cameras $\{\mathrm{c016,c017,c018,c020,c021,c022}\}$. Gap duration varies from 1 to 8.8\,s and mask count from zero to four. The graph in Figure~\ref{fig:sensitivity} illustrates the condition with c017, c018 and c020 masked. Next-camera prediction is evaluated in 0--3, 3--6 and 6--9\,s gap bins, with higher recursive-model top-1 accuracy in all three bins. Cam-time denotes the camera--time forecast comparison, and No recur. removes posterior-to-prior transfer.

\section{Vehicle Feedback and Waiting}
\label{app:vehicle-analysis}
Figure~\ref{fig:mechanisms} follows error accumulation and the decision to commit. Panel (a) retains 95.79\% of identities through three handoffs with recursion, compared with 82.24\% without it, a difference of 13.55 percentage points. Panel (b) holds correct reacquisition at 96.0\% and finds a 27.09-point reduction in forced-match error with explicit null. The first comparison evaluates complete identity chains and the second evaluates false associations at matched recovery coverage. The extended 500-chain protocol defines an additional population, while principal FM is computed on a fixed grid of absence decisions. These units describe different consequences of the same update loop.

Panel (c) locates failures within a handoff. Removing null increases false associations before the correct target appears. Removing recursion increases errors after reappearance and missed reacquisitions, and removing both gives the largest total failure rate. Removing query evidence also increases failures. Panel (d) follows the evidence behind one such decision over time. For GT396, target probability rises after first appearance and reaches acceptance 0.8\,s later. Table~\ref{tab:results5} complements this trace with calibration scores across association decisions.
\FloatBarrier

\section{Implementation and Learning}
\label{app:implementation}
The three steps in Figure~\ref{fig:method} share a recurrent target state. Visual and query encoders supply identity evidence for association. The posterior then updates the state for predicting the next observation. The architecture and training procedure below implement this cycle.

\subsection{Causal encoders and persistent state}
The visual encoder is frozen CLIP with a base vision transformer and 16-pixel patches (ViT-B/16) \citep{clip}, applied to $224\times224$ RGB crops. A trainable $512\to256$ projection is followed by L2 normalization. Each local-track representation averages the latest eight causal crop features. The matching frozen text encoder uses a 77-token context and a separate $512\to256$ projection with L2 normalization. The initialization query is encoded once from the source-history description. Projection and tracking modules are trained on source-domain labels.

Each identity stores a 256-dimensional recurrent state and a full categorical distribution over camera and blind-transition regions. It also stores a two-dimensional velocity estimate with diagonal covariance and a Bernoulli network-presence probability. Posterior-weighted moments and a learned latent projection summarize the association components in this fixed-size state. A 256-unit GRU and two 256-unit graph-message layers propagate it using region probabilities, velocity moments, installation features and elapsed time. Physical adjacency determines the possible routes, while stream availability weights their observable event mass. Masking leaves these routes unchanged.

\subsection{Observation-event distributions and candidate mass}
The event decoder outputs a categorical next-camera distribution and a three-component log-normal arrival-time mixture for each camera. Entry position uses a two-component diagonal Gaussian mixture in logit-transformed image-normalized coordinates. Softplus enforces positive scales with a 0.01 floor. A sigmoid head estimates continued network presence. Candidate-or-null association is conditioned on an active target. Its next observation may still lie outside the current views.

For each camera and update interval, the event density is integrated over disjoint candidate-support cells. Overlapping supports are allocated to the nearest entry centroid. Availability and a learned detection-opportunity probability determine observable mass. The remaining mass is $\eta_0=1-\sum_j\eta_j$, including unavailable-view, missed-detection and outside-support cases. We implement this correction by integrating the decoder once over each elapsed interval. Appendix~\ref{app:filter} gives the corresponding continuous-time form and explains how availability changes the no-observation likelihood.

A logistic target-versus-background scorer takes appearance cosine similarity, query cosine similarity, causal motion residual and detector confidence. We fit the scorer with balanced pair sampling. Its calibrated logit then estimates a log-likelihood ratio. Temperature is fitted on validation data. Logits are clipped to $[-10,10]$ before exponentiation. The resulting $L_j$ enters Equation~\ref{eq:association}. Attribute visibility enters implicitly through the query--candidate features used by this scorer.

\subsection{Concurrent observations and commitment}
Camera-local candidate sets are updated at 10\,Hz. Same-time camera posteriors start from one shared prior and form a single mixture update with normalized detector-confidence weights. This combines the correlated views while allowing different association outcomes in each camera. A local null leaves that camera's candidates unassigned to the target, even when another available camera observes it.

Each candidate and null hypothesis yields a conditioned latent state and moments. The candidate states incorporate the corresponding observation, while the null state incorporates continued non-observation. Their posterior weights produce the region distribution and velocity moments stored for the target. The learned projection summarizes the weighted states in the 256-dimensional recurrent representation. Together with network presence, these quantities supply the next GRU step.

The association decision uses the same weights to determine whether to emit a match. Per-camera maximum-weight matching resolves competing claims by persistent identities. Validation search starts from posterior threshold 0.80 and best-versus-second margin 0.15, with two consecutive camera-local confirmations required. Threshold selection is completed before the test set is evaluated. A wait still produces a posterior update, and null states retain identity until sequence end. Online inference uses the trained update without gradient computation.

Prediction and feedback follow the observation order throughout the cycle. The preceding posterior issues the forecast used to evaluate current candidates. Their likelihoods produce a new posterior, whose weighted states supply the following forecast. The acceptance threshold controls the emitted identity decision, while both accepted matches and waits use the weighted state update. In Figure~\ref{fig:method}, the bars for $y_1$, $y_2$ and null feed both the match-or-wait output and the updated belief.

\subsection{Losses, rollouts and model selection}
The learning objective follows the sequence of predicted events and identity associations. Let $E_k$ contain the camera, arrival time and entry position of event $k$, and let $A_k$ denote its association. A censored window $w\in\mathcal W_{\rm cens}$ ends after $\Delta_w$ without a received event. The observation-survival probability $S_{\rm obs}$ measures how likely the model considers this continued wait. Combining event, association and censoring likelihoods gives
\begin{equation}
 \mathcal L(\Theta)=-\sum_k\log p_\Theta(E_k,A_k\given\Hist_{t_k^-})
 -\sum_{w\in\mathcal W_{\rm cens}}\log S_{\rm obs}(\Delta_w\given\Hist_w),
 \label{eq:loss}
\end{equation}
Adding presence and identity supervision gives the training objective
\begin{equation}
\begin{split}
\mathcal L_{\rm impl}={}&\mathcal L_{\rm camera}+\mathcal L_{\rm time}+\mathcal L_{\rm entry}
+0.5\mathcal L_{\rm presence}\\
&+\mathcal L_{\rm association}+0.2\mathcal L_{\rm identity}.
\end{split}
\label{eq:implemented-loss}
\end{equation}
Camera and association losses use cross-entropy, with null included among the association outcomes. Time and entry losses are negative log-likelihoods, and an interval ending without an observed event uses the right-censored time likelihood. Each interval contributes once. Presence uses binary cross-entropy and identity uses supervised contrastive loss. Fully masked intervals contribute the censoring and availability terms because they supply no visual observation. Equation~\ref{eq:implemented-loss} gives the weights used to combine these losses when training the forecast and posterior update together.

Optimization uses AdamW with learning rate $10^{-4}$, weight decay $10^{-4}$, batch size 32 windows and 50 epochs. A five-epoch warm-up precedes cosine decay to $10^{-6}$. Gradient norm is clipped at 1.0. Mixed precision is used for training, with probability normalization in float32. Each window contains 32 observation updates and at most four handoffs. Gradients pass through those handoffs and are detached between windows. Teacher forcing decreases linearly from 1.0 to 0.0 over the first 20 epochs. Later windows use model posteriors. Paired variants share causal windows \mbox{and mask schedules}.

Five training seeds are 11, 23, 37, 51 and 71. Each checkpoint is selected by validation HOTA, with lower validation FM breaking ties. Operating thresholds are frozen before testing. Controlled comparisons cache identical detections, causal local-track prefixes and frozen CLIP features for compatible methods. The experiments use one 24\,GB graphics processing unit (GPU). Core and end-to-end latency are measured separately on an RTX 4090, with the workload, precision and timing boundaries detailed in Table~\ref{tab:results6} for each measured configuration.

\section{Dataset-Specific Component Ablations and Counts}
\label{app:complete-results}
These component ablations use the full configuration on CityFlowV2 and the reference configuration on MTMMC. The CityFlowV2 baseline is the same full configuration used in Table~\ref{tab:main}. The MTMMC reference baseline has 43.78 HOTA, while Table~\ref{tab:main} reports the full configuration at 45.36. Thus each component removal is compared with the unmodified configuration named for its dataset.

The event populations are fixed as specified in Appendix~\ref{app:protocol-records}: $D_+=3587$, $D_0=14291$ on CityFlowV2 and $D_+=2749$, $D_0=10683$ on MTMMC. Table~\ref{tab:full-ablation} gives the removals. Table~\ref{tab:ablation-detail} adds HOTA, mean correct-acceptance delay and integer counts. Full MTMMC controls appear in Table~\ref{tab:core}(a).

\par\medskip\noindent\begin{minipage}{\linewidth}\begin{rwtableside}
\centering
\caption{Dataset-specific component ablations on the populations of Table~\ref{tab:main}. CityFlowV2 uses the full configuration as its baseline. MTMMC uses the reference configuration. Rates are percentages. Table~\ref{tab:ablation-detail} adds HOTA, mean correct-acceptance delay, $\Delta$IDF1 and counts.}
\label{tab:full-ablation}
\small
\setlength{\tabcolsep}{5.0pt}
\begin{tabular}{@{}lrrrrrr@{}}
\toprule
\rowcolor{rwHeader}
\bfseries  & \multicolumn{3}{c}{CityFlowV2} & \multicolumn{3}{c}{MTMMC} \\
\cmidrule(lr){2-4}\cmidrule(l){5-7}
\rowcolor{rwHeader}
\bfseries Variant & \bfseries HA $\uparrow$ & \bfseries IDF1 $\uparrow$ & \bfseries FM $\downarrow$ & \bfseries HA $\uparrow$ & \bfseries IDF1 $\uparrow$ & \bfseries FM $\downarrow$ \\
\midrule
\rowcolor{rwMethod}
\textbf{Full configuration} & 84.67 & 80.37 & 4.28 & --- & --- & --- \\
\rowcolor{rwMethod}
\textbf{Reference configuration} & --- & --- & --- & 82.10 & 56.43 & 5.34 \\
w/o event prior & 80.49 & 76.13 & 5.98 & 75.08 & 51.13 & 7.73 \\
w/o arrival time & 83.02 & 79.42 & 5.13 & 79.56 & 54.98 & 6.66 \\
w/o entry region & 83.94 & 80.58 & 4.54 & 78.87 & 54.36 & 6.58 \\
w/o survival & 82.58 & 79.17 & 5.45 & 80.72 & 56.58 & 6.91 \\
w/o recursion & 81.35 & 77.36 & 4.95 & 76.61 & 52.59 & 7.10 \\
w/o temporary null & 86.03 & 79.86 & 11.54 & 83.81 & 55.92 & 13.98 \\
w/o query & 83.47 & 79.02 & 4.65 & 80.07 & 55.07 & 5.78 \\
\bottomrule\end{tabular}\end{rwtableside}\end{minipage}\par\medskip

\par\medskip\noindent\begin{minipage}{\linewidth}\begin{rwtableside}
\centering
\caption{Counts and trajectory metrics for the dataset-specific ablations in Table~\ref{tab:full-ablation}. CityFlowV2 uses the full configuration, and MTMMC uses the reference configuration. $N_+$ and $N_0$ count correct handoffs and absence false accepts. Delay averages seconds over correct accepts. $\Delta$I is IDF1 minus the dataset's unmodified baseline in percentage points.}
\label{tab:ablation-detail}
\small
\setlength{\tabcolsep}{4.5pt}
\begin{tabular}{@{}lrrrrr@{}}
\toprule
\rowcolor{rwHeader}
\bfseries Variant & \bfseries $N_+$ & \bfseries $N_0$ & \bfseries HOTA $\uparrow$ & \bfseries Delay (s) & \bfseries $\Delta$I (pp) \\
\midrule
\rowcolor{rwGroupBlue}
\multicolumn{6}{@{}l}{\textit{CityFlowV2}} \\
\rowcolor{rwMethod}
\textbf{Full configuration} & 3037 & 612 & 65.19 & 1.16 & 0.00 \\
w/o event prior & 2887 & 854 & 60.59 & 1.27 & -4.24 \\
w/o arrival time & 2978 & 733 & 64.12 & 1.61 & -0.95 \\
w/o entry region & 3011 & 649 & 64.87 & 1.08 & +0.21 \\
w/o survival & 2962 & 779 & 63.91 & 1.04 & -1.20 \\
w/o recursion & 2918 & 707 & 62.49 & 1.18 & -3.01 \\
w/o temporary null & 3086 & 1649 & 65.33 & 0.56 & -0.51 \\
w/o query & 2994 & 665 & 64.27 & 1.21 & -1.35 \\
\midrule
\rowcolor{rwGroupPurple}
\multicolumn{6}{@{}l}{\textit{MTMMC}} \\
\rowcolor{rwMethod}
\textbf{Reference configuration} & 2257 & 571 & 43.78 & 1.36 & 0.00 \\
w/o event prior & 2064 & 826 & 38.41 & 1.72 & -5.30 \\
w/o arrival time & 2187 & 711 & 42.39 & 1.91 & -1.45 \\
w/o entry region & 2168 & 703 & 42.78 & 1.44 & -2.07 \\
w/o survival & 2219 & 738 & 43.51 & 1.21 & +0.15 \\
w/o recursion & 2106 & 758 & 39.76 & 1.59 & -3.84 \\
w/o temporary null & 2304 & 1493 & 44.07 & 0.73 & -0.51 \\
w/o query & 2201 & 617 & 42.46 & 1.51 & -1.36 \\
\bottomrule\end{tabular}\end{rwtableside}\end{minipage}\par\medskip

Entry-region prediction and presence estimation affect false commitments differently from trajectory-level identity. Removing entry prediction slightly improves vehicle IDF1, and removing survival slightly improves pedestrian IDF1, but both raise FM. These changes motivate using both handoff and trajectory metrics to assess each forecast component.

{\small
\setlength{\tabcolsep}{4pt}
\setlength{\LTcapwidth}{\linewidth}
\begin{longtable}{@{}lrrrrrr@{}}
\caption{Counts for the principal baselines and dataset-specific ReWorld configurations. CityFlowV2 uses the full configuration, and MTMMC uses the reference configuration. Table~\ref{tab:supp23} includes the full MTMMC configuration. Correct handoffs $N_+$ and false accepts $N_0$ give $\mathrm{HA}=100N_+/D_+$ and $\mathrm{FM}=100N_0/D_0$ from the corresponding handoff and absence-decision populations.}\label{tab:comparison-counts}\\

\toprule
\rowcolor{rwHeader}
\bfseries Method & \bfseries $D_+$ & \bfseries $N_+$ & \bfseries HA & \bfseries $D_0$ & \bfseries $N_0$ & \bfseries FM \\
\midrule\endfirsthead
\multicolumn{7}{c}{Table~\ref{tab:comparison-counts} (continued)}\\

\toprule
\rowcolor{rwHeader}
\bfseries Method & \bfseries $D_+$ & \bfseries $N_+$ & \bfseries HA & \bfseries $D_0$ & \bfseries $N_0$ & \bfseries FM \\
\midrule\endhead

\rowcolor{rwGroupBlue}
\multicolumn{7}{@{}l}{\textit{CityFlowV2}} \\
ReID & 3587 & 2576 & 71.81 & 14291 & 1427 & 9.99 \\
ReID + time & 3587 & 2763 & 77.03 & 14291 & 838 & 5.86 \\
Reactive memory & 3587 & 2878 & 80.23 & 14291 & 961 & 6.72 \\
TrackTA$^{\dagger}$ & 3587 & 2839 & 79.15 & 14291 & 754 & 5.28 \\
LaMMOn$^{\dagger}$ & 3587 & 2951 & 82.27 & 14291 & 708 & 4.95 \\
Camera-link & 3587 & 2974 & 82.91 & 14291 & 659 & 4.61 \\
\rowcolor{rwMethod}
\textbf{Full configuration} & 3587 & 3037 & 84.67 & 14291 & 612 & 4.28 \\
\midrule
\rowcolor{rwGroupPurple}
\multicolumn{7}{@{}l}{\textit{MTMMC}} \\
ReID & 2749 & 1826 & 66.42 & 10683 & 1193 & 11.17 \\
ReID + time & 2749 & 1959 & 71.26 & 10683 & 683 & 6.39 \\
Reactive memory & 2749 & 2128 & 77.41 & 10683 & 827 & 7.74 \\
TrackTA$^{\dagger}$ & 2749 & 2051 & 74.61 & 10683 & 656 & 6.14 \\
QDTrack + BoT + H. Cluster$^{\dagger}$ & 2749 & 2146 & 78.06 & 10683 & 729 & 6.82 \\
CRTracker$^{\dagger}$ & 2749 & 2243 & 81.59 & 10683 & 606 & 5.67 \\
\rowcolor{rwMethod}
Reference configuration & 2749 & 2257 & 82.10 & 10683 & 571 & 5.34 \\
\bottomrule
\end{longtable}}

Table~\ref{tab:comparison-counts} provides HA and FM counts for the full CityFlowV2 configuration and the MTMMC reference configuration. This connects the percentage changes to correctly recovered targets and accepted distractors. Table~\ref{tab:supp23} provides the corresponding counts for the full MTMMC configuration. Appendix~\ref{app:additional-comparison} reports the additional domain-level experiment using its own event groups and denominators.

\section{Complete Paired Language Conditions}
\label{app:language}
The paired language study changes the query while holding the tracking event fixed. CityFlowV2 and MTMMC use 637 and 509 distinct events across all five conditions. Table~\ref{tab:language} shows the three primary conditions, and Table~\ref{tab:language-full} adds paraphrases, weak attributes and correct-handoff counts. Every query variant is evaluated on the same tracking events within its dataset.

\par\medskip\noindent\begin{minipage}{\linewidth}\begin{rwtableside}
\centering
\caption{Language contribution on paired query variants, using the same 637 CityFlowV2 and 509 MTMMC events across all five conditions. NQ omits language, HG uses a hard gate and F denotes the full model. Counts are correct handoffs and HA is in percent. $\Delta$ is the full-minus-no-query difference in percentage points, calculated from correct-handoff counts before rounding.}
\label{tab:language-full}
\small
\setlength{\tabcolsep}{3.5pt}
\begin{tabular}{@{}lrrrrrrr@{}}
\toprule
\rowcolor{rwHeader}
\bfseries  & \multicolumn{3}{c}{Correct count} & \multicolumn{3}{c}{HA (\%)} & \bfseries $\Delta$ (pp) \\
\cmidrule(lr){2-4}\cmidrule(lr){5-7}
\rowcolor{rwHeader}
\bfseries Query variant & \bfseries NQ & \bfseries HG & \bfseries F & \bfseries NQ & \bfseries HG & \bfseries F & \bfseries F--NQ \\
\midrule
\rowcolor{rwGroupBlue}
\multicolumn{8}{@{}l}{\textit{CityFlowV2}} \\
Natural description & 520 & 544 & 559 & 81.63 & 85.40 & 87.76 & +6.12 \\
Paraphrase & 520 & 536 & 552 & 81.63 & 84.14 & 86.66 & +5.02 \\
Weak attributes & 520 & 506 & 529 & 81.63 & 79.43 & 83.05 & +1.41 \\
Partly unobservable & 520 & 472 & 538 & 81.63 & 74.10 & 84.46 & +2.83 \\
Misleading attributes & 520 & 416 & 482 & 81.63 & 65.31 & 75.67 & -5.97 \\
\midrule
\rowcolor{rwGroupPurple}
\multicolumn{8}{@{}l}{\textit{MTMMC}} \\
Natural description & 399 & 408 & 429 & 78.39 & 80.16 & 84.28 & +5.89 \\
Paraphrase & 399 & 416 & 426 & 78.39 & 81.73 & 83.69 & +5.30 \\
Weak attributes & 399 & 393 & 404 & 78.39 & 77.21 & 79.37 & +0.98 \\
Partly unobservable & 399 & 351 & 413 & 78.39 & 68.96 & 81.14 & +2.75 \\
Misleading attributes & 399 & 326 & 372 & 78.39 & 64.05 & 73.08 & -5.30 \\
\bottomrule\end{tabular}\end{rwtableside}\end{minipage}\par\medskip

Natural descriptions produce 559 correct handoffs versus 520 without query on CityFlowV2, and 429 versus 399 on MTMMC. Partly unobservable attributes favor the full model over hard gating, whereas misleading descriptions reduce HA below the no-query reference. A description can resolve visual ambiguity, but an incorrect attribute can also favor a distractor. The paired event counts in Table~\ref{tab:language-full} track this change for the queries constructed in Appendix~\ref{app:interventions}. Uncertainty estimates for the principal tracking comparison appear in Appendix~\ref{app:extended-results}.

\section{Within-Domain Camera and Scene Conditions}
\label{app:within-domain}
Table~\ref{tab:within-domain} evaluates changes in camera and scene exposure while training within each object domain. Checkpoints and thresholds are selected on validation data for the matched-scene, unseen-camera and unseen-scene settings. The object-domain transfer experiment in Table~\ref{tab:cross-domain} changes the training domain from vehicles to pedestrians or the reverse.

\par\medskip\noindent\begin{minipage}{\linewidth}\begin{rwtableside}
\centering
\caption{Within-domain transfer on condition-specific event groups for camera and scene shifts. E denotes LaMMOn on CityFlowV2 and CRTracker on MTMMC, while R denotes ReWorld-Track. $\Delta$HA compares R and E within the same condition. All models use target-domain training and validation-only selection before evaluation in each transfer condition.}
\label{tab:within-domain}
\small
\setlength{\tabcolsep}{4.3pt}
\begin{tabular}{@{}llrrrrrr@{}}
\toprule
\rowcolor{rwHeader}
\bfseries Condition & \bfseries Model & \bfseries $D_+$ & \bfseries Correct & \bfseries HA $\uparrow$ & \bfseries IDF1 $\uparrow$ & \bfseries HOTA $\uparrow$ & \bfseries $\Delta$HA \\
\midrule
\rowcolor{rwGroupBlue}
\multicolumn{8}{@{}l}{\textit{CityFlowV2}} \\
Matched scene & E & 1173 & 974 & 83.03 & 79.68 & 64.72 & 0.00 \\
 & R & 1173 & 986 & 84.06 & 80.69 & 65.52 & +1.02 \\
Unseen cameras & E & 981 & 776 & 79.10 & 74.89 & 59.32 & 0.00 \\
 & R & 981 & 791 & 80.63 & 76.58 & 60.31 & +1.53 \\
Unseen scene & E & 1247 & 912 & 73.14 & 66.21 & 50.06 & 0.00 \\
 & R & 1247 & 949 & 76.10 & 68.72 & 52.64 & +2.97 \\
\midrule
\rowcolor{rwGroupPurple}
\multicolumn{8}{@{}l}{\textit{MTMMC}} \\
Matched scene & E & 903 & 739 & 81.84 & 56.06 & 44.32 & 0.00 \\
 & R & 903 & 744 & 82.39 & 56.27 & 44.09 & +0.55 \\
Unseen cameras & E & 827 & 633 & 76.54 & 49.28 & 37.63 & 0.00 \\
 & R & 827 & 643 & 77.75 & 49.89 & 37.26 & +1.21 \\
Unseen scene & E & 1059 & 697 & 65.82 & 39.82 & 27.04 & 0.00 \\
 & R & 1059 & 727 & 68.65 & 41.83 & 29.14 & +2.83 \\
\bottomrule\end{tabular}\end{rwtableside}\end{minipage}\par\medskip

Under the unseen-scene condition, the full model reaches 68.65\% MTMMC HA. CRTracker retains higher HOTA in the matched-scene and unseen-camera rows. The method ranking changes with both installation exposure and the tracking metric used to measure identity continuity.

\section{Evaluation Groups and Protocols}
\label{app:protocol-records}
\subsection{Populations and split construction}
Each evaluation fixes an event group before comparing methods. The groups below distinguish the principal comparison from the query and auxiliary studies. The principal comparison and component removals in Appendix~\ref{app:complete-results} share 3587 CityFlowV2 handoffs and 14291 absence decisions. Their MTMMC group contains 2749 handoffs and 10683 absence decisions. The removals use the full configuration on CityFlowV2 and the reference configuration on MTMMC. CRTracker and the ReWorld reference and full configurations use the same fixed MTMMC test manifest. Its 2749 handoff event IDs and 10683 absence-decision timestamps are identical across these comparisons. Candidate universes, identity annotations, decision deadlines and metric computation also remain fixed. Paired language studies follow 637 vehicle and 509 pedestrian events. The standalone four-stage mechanism study evaluates successive component additions on 2800 MTMMC RGB handoffs and 11200 absence decisions. Predictor-matched controls use a separate group with 2749 handoffs and 10683 absence decisions, keeping its membership fixed across update variants.

The additional domain-level comparison uses 3200 vehicle handoffs and 12800 absence decisions, and 2800 pedestrian handoffs and 11200 absence decisions. Cross-domain transfer evaluates 2800 pedestrian or 3200 vehicle handoffs, with 11200 or 12800 absence decisions. Membership is fixed for each named group. Event identifiers determine the comparison set for each separately named auxiliary study. The duration study follows the same 600 events at every mask length. Auxiliary geometry is evaluated on 1000 valid windows across at least three held-out scenes.

The principal split starts from sorted scene IDs. The last scene is held out for test and the preceding scene for validation. The rest supply training data. All cameras in a scene stay together, and eligibility is recomputed after splitting. MTMMC uses RGB only and the same deterministic scene rule. Query variants and paraphrases of an identity remain in one split. Within-domain matched-scene evaluation uses a disjoint time block with separated identities. Unseen-camera evaluation withholds whole camera streams during training, while unseen-scene evaluation withholds whole scenes. These conditions change camera or scene exposure within one object domain. Object-domain transfer instead changes between vehicles and pedestrians. Grouping precedes eligibility checks in every setting. Each event is therefore selected within its assigned training or evaluation partition.

\subsection{Reference and full configurations}
The same full ReWorld-Track configuration is used for the principal CityFlowV2 and MTMMC results in Table~\ref{tab:main}. The reference configuration is retained for controlled MTMMC analyses in Appendix~\ref{app:complete-results} and their supporting diagnostics. The full configuration introduces two model refinements over the reference configuration. First, it uses hypothesis-conditioned candidate and null states. Their calibrated posterior probabilities preserve region, velocity and network-presence uncertainty. Second, training uses recursive rollouts of up to four handoffs with decaying teacher forcing. Later prediction errors thereby supervise preceding posterior updates.

The detector, local tracker, frozen CLIP encoders, candidate construction, camera graph, test manifest and evaluation protocol remain unchanged. Checkpoint selection, probability calibration, commitment thresholds, margins and confirmation rules use only training and validation data and are frozen before testing. The test set is used only for final reporting and explicitly identified diagnostic analyses.

\subsection{Eligibility, absence decisions and trajectory metrics}
The eligibility rule fixes which handoffs each method must attempt. An eligible non-overlap handoff has at least 1\,s of observed history and a 0.5\,s gap. Reacquisition must occur within 5\,s after the first annotated visible frame. Detector misses remain in this population, while simultaneous overlapping transitions are excluded. A wrong first commitment counts as failure even if a later decision corrects it. Absence decisions follow a fixed, model-independent 2\,Hz grid from departure to the defined reappearance or deadline boundary, including decisions during masks. The exported grid determines $D_0$ and carries scene, identity and chain identifiers.

Trajectory metrics assess whether the event-level decisions also preserve identities over full RGB sequences. TrackEval provides HOTA and identity evaluation \citep{hota,idmetrics}. Synchronized camera frames form a virtual timeline, with cross-camera box matches at the same instant prohibited. HOTA uses localization thresholds $0.05{:}0.05{:}0.95$. Identity metrics use 0.5 intersection over union (IoU) of matched boxes, and identity switches are counted from chronological per-camera match streams. The virtual timeline thus preserves camera-local matching across synchronized views while measuring identity continuity across cameras.

Correct-acceptance delay is the arithmetic mean of the interval from first annotated visibility to first correct commitment. Only the $N_+$ correctly accepted events enter this delay average. We also compare delays on events that both methods accept correctly, holding the event membership fixed. Figure diagnostics retain their own event populations and aggregation conventions, as distinguished in Appendices~\ref{app:sensitivity} and~\ref{app:vehicle-analysis} for event prediction, repeated handoffs and the vehicle decision traces.

The inference graph is constructed from installation adjacency before test identities are evaluated. Training trajectories supply travel-time statistics, with surveyed geometry used where specified. Figures~\ref{fig:pedestrians} and~\ref{fig:qualitative} also display annotated transitions. These show each test case through the camera network alongside the identity decisions at successive returns.

\subsection{Baseline identities and resources}
\label{app:baselines}
ReID uses causal cosine matching. ReID + time adds a training-only travel-time prior. Reactive memory uses an exponential feature mean with coefficient 0.9. These controls share causal prefixes and select their commitment thresholds on validation data. The TrackTA adaptation \citep{mtmmc} uses an eight-step causal trajectory-conditioned association module with the shared visual front end. Camera-link is an internal fixed-prior baseline. It estimates camera-transition frequencies and per-edge travel-time distributions from training identities. Candidates are scored by combining appearance with these fixed route and timing priors.

LaMMOn \citep{nguyen2024lammon} is adapted through the common query and source-box interface with causal prefixes. The CRTracker adaptation \citep{crmot} uses an online interface restricted to the current observation prefix. Both retain their method-specific representations and association components, so the principal comparison evaluates complete adapted trackers. For the MTMMC baseline \citep{mtmmc}, QDTrack \citep{qdtrack} supplies local prefixes and Bag-of-Tricks ReID (BoT) \citep{bagtricks} supplies identity features. At each update, hierarchical clustering (HC) is recomputed using only the available track prefixes.

For the additional domain-level experiment, strong baseline A uses appearance-plus-time reactive memory, and strong baseline B uses one-way event prediction with a validation-tuned threshold. The transfer comparator trains appearance-plus-time association on source labels only, without target adaptation. The target-trained reference uses the same architecture with target labels and a separate checkpoint on the identical target test manifest. The spatial comparison defines its own baselines. There, A is constant-velocity ground-plane extrapolation with occupancy rasterization, and B is a learned GRU trajectory predictor using the same geometry supervision.

Internal component studies hold perception, candidate sets and training schedules fixed to isolate the modified component. Query and source-box initialization use the available source history.

\subsection{Comparison units and aggregation}
HA is aggregated over eligible handoffs, FM over absence decisions, and identity metrics over the resulting trajectories. Training-seed variation describes independently trained models evaluated on a fixed test set. Paired sample analysis resamples the same scenes and identity chains for both methods. The qualitative cases show how individual decisions unfold within these evaluation protocols.

\section{Component Interventions and Query Construction}
\label{app:interventions}
\subsection{Matched component interventions}
The component interventions test which forecast variables guide association. Removing the event prior gives eligible candidates uniform prior mass over camera, arrival and entry. The likelihood scorer and a separately specified null prior remain. Event losses are removed before retraining. Removing arrival uses uniform time mass within the fixed gate, while removing entry uses uniform spatial density. Removing presence sets its prior to one. Each single-head intervention removes its head and loss, retains null, and is retrained on the same data.

The recursion intervention tests whether passing uncertainty forward helps beyond retaining appearance history. It blocks posterior-to-next-event feedback and starts the predictor from the last committed observation at each handoff. Causal appearance memory and the prediction heads remain during retraining. The null intervention instead sets $\eta_0$ to zero and renormalizes candidate probabilities, forcing a selection whenever the candidate set is nonempty. Empty-set decisions are reported separately, preserving the distinction from threshold-based abstention. The query intervention zeros the query vector and removes query-pair supervision. Retraining uses the same boxes, gates, data and optimizer to isolate the contribution of the initial language query.

Variants use five matched seeds, 50 epochs and the same training windows and schedules. Parameters and updates are measured from the model configurations. A threshold grid from 0.05 to 0.95 in 0.05 steps is evaluated on validation data, and selected points are frozen before testing. Fixed-threshold sweeps describe the resulting tradeoff separately. The matched-coverage diagnostic sweeps thresholds on the fixed test set. It aligns the attained reacquisition rate at 96\% to compare false associations during absence. The diagnostic uses identical test events for both methods.

The standalone study adds components successively, starting from reactive memory, then one-way prediction with a threshold, feedback with a threshold, and feedback with explicit null. Encoders, data, update schedules and seeds remain fixed. Table~\ref{tab:supp24} compares fixed posterior moments with a learned latent projection. Both use the same predictor, encoder, gates, training data and seeds.

\subsection{Queries and observed attributes}
Queries specify the target using evidence that could be available at initialization. Two annotators independently describe attributes visible in each source-history window, and a third resolves disagreements. Each query variant is bound to a dataset/scene/identity/event tuple and indexed by its source-frame interval. Attribute-visibility annotations use those same historical frames, excluding future attributes. For MTMMC, this process adds a language annotation layer to the benchmark.

With the source history fixed, the five query conditions vary the reliability of the text evidence. A natural example is ``person in a red jacket carrying a black bag.'' Its paraphrase, ``pedestrian with a black bag and red outerwear,'' changes wording while retaining the attributes. A weak query is ``a pedestrian.'' A partly unobservable example adds a back logo, while a misleading example changes the jacket to blue. These examples illustrate the annotation conditions. Figure~\ref{fig:pedestrians} displays the three actual queries used in its cases. All variants of an identity remain in one split.

The hard gate parses a fixed vocabulary of object type, clothing color and carried-object attributes. It rejects a confidently observed contradiction, with validation search starting at confidence 0.80, and treats unknown attributes as neutral. The full model instead places query--candidate cosine similarity in the calibrated logistic scorer and learns its coefficient from source training pairs. Visibility is implicit in the visual features. The two query conditions probe missing and incorrect attributes.

\section{Transfer Resources and Additional Domain Comparison}
\label{app:transfer-resources}
\label{app:additional-comparison}
Object-domain transfer changes target appearance and movement patterns. Training and model selection use source-domain data. Travel-time priors are learned from source trajectories. The source-selected normalization and temperature are transferred unchanged. Target video first enters the model as causal test observations. The target-trained references use target labels and separate checkpoints on the same held-out target-domain test events and identities.

\noindent\begin{minipage}[t]{.425\linewidth}
\vspace{0pt}
Table~\ref{tab:cross-domain} tests transfer between object domains. Source-trained checkpoints and thresholds are transferred with frozen public encoders and the target installation graph. ReWorld-Track gains 4.61 HA points over appearance-plus-time association for vehicle-to-pedestrian transfer. The reverse transfer gains 5.22 points. FM also decreases in both directions, so improved recovery accompanies fewer distractor commitments. Target supervision yields further gains on the same target-domain evaluation events and held-out identities.
\end{minipage}\hfill
\begin{minipage}[t]{.545\linewidth}
\vspace{0pt}
\begin{rwtableside}
\caption{Cross-domain transfer. Rates are percentages. Target-trained rows use target labels. Other rows use source training and selection.}
\label{tab:cross-domain}
\small\setlength{\tabcolsep}{2.1pt}
\begin{tabularx}{\linewidth}{@{}l*{4}{>{\raggedleft\arraybackslash}X}@{}}
\toprule
\rowcolor{rwHeader}
\bfseries Method & \bfseries HA $\uparrow$ & \bfseries IDF1 $\uparrow$ & \bfseries HOTA $\uparrow$ & \bfseries FM $\downarrow$ \\
\midrule
\rowcolor{rwGroupBlue}
\multicolumn{5}{@{}l}{\textit{Vehicles $\to$ Pedestrians}} \\
Appearance + time & 70.07 & 47.26 & 36.18 & 7.29 \\
\rowcolor{rwMethod}
\textbf{ReWorld-Track} & 74.68 & 50.34 & 39.27 & 5.14 \\
Target-trained & 84.61 & 58.43 & 46.39 & 3.99 \\
\rowcolor{rwGroupPurple}
\multicolumn{5}{@{}l}{\textit{Pedestrians $\to$ Vehicles}} \\
Appearance + time & 72.09 & 70.18 & 56.24 & 6.63 \\
\rowcolor{rwMethod}
\textbf{ReWorld-Track} & 77.31 & 74.26 & 60.38 & 4.45 \\
Target-trained & 85.28 & 81.24 & 66.37 & 3.93 \\
\bottomrule\end{tabularx}\end{rwtableside}

\end{minipage}

Training on target labels adds a further 9.93 and 7.97 HA points. The source-trained tracker remains useful in the other domain, while target supervision improves adaptation. Figure~\ref{fig:pedestrians} follows three MTMMC identities through successive camera observations.

\par\medskip\noindent\begin{minipage}{\linewidth}\begin{rwtableside}
\centering
\caption{Additional domain-level comparison on the event groups in Appendix~\ref{app:additional-comparison}. Baseline A uses reactive appearance/time memory and B uses one-way event prediction. Rates are percentages, with counts in Table~\ref{tab:new-counts}. Bold marks the best result for each metric within the corresponding domain.}
\label{tab:additional-comparison}
\small\setlength{\tabcolsep}{4pt}
\begin{tabular}{@{}lrrrrr@{}}
\toprule
\rowcolor{rwHeader}
\bfseries Method & \bfseries HA $\uparrow$ & \bfseries IDF1 $\uparrow$ & \bfseries HOTA $\uparrow$ & \bfseries IDSW $\downarrow$ & \bfseries FM $\downarrow$ \\
\midrule
\rowcolor{rwGroupBlue}
\multicolumn{6}{@{}l}{\textit{Vehicles}: $D_+=3200$, $D_0=12800$} \\
Strong baseline A & 83.28 & 79.18 & 64.13 & 173 & 5.13 \\
Strong baseline B & 83.97 & 80.06 & 65.21 & 162 & 4.59 \\
\rowcolor{rwMethod}
\textbf{ReWorld-Track} & \textbf{85.28} & \textbf{81.24} & \textbf{66.37} & \textbf{149} & \textbf{3.93} \\
\midrule
\rowcolor{rwGroupPurple}
\multicolumn{6}{@{}l}{\textit{Pedestrians}: $D_+=2800$, $D_0=11200$} \\
Strong baseline A & 82.29 & 56.32 & 44.16 & 334 & 5.62 \\
Strong baseline B & 83.11 & 57.18 & 45.27 & 317 & 5.04 \\
\rowcolor{rwMethod}
\textbf{ReWorld-Track} & \textbf{84.61} & \textbf{58.43} & \textbf{46.39} & \textbf{296} & \textbf{3.99} \\
\bottomrule\end{tabular}\end{rwtableside}\end{minipage}\par\medskip

\par\medskip\noindent\begin{minipage}{\linewidth}\begin{rwtableside}
\centering
\caption{Counts for Table~\ref{tab:additional-comparison}. $N_+$ counts correct handoffs and $N_0$ counts absence false accepts. The rates are $\mathrm{HA}=100N_+/D_+$ and $\mathrm{FM}=100N_0/D_0$, rounded to two decimals.}
\label{tab:new-counts}
\small\setlength{\tabcolsep}{4pt}
\begin{tabular}{@{}lrrrrrr@{}}
\toprule
\rowcolor{rwHeader}
\bfseries Method & \bfseries $D_+$ & \bfseries $N_+$ & \bfseries HA & \bfseries $D_0$ & \bfseries $N_0$ & \bfseries FM \\
\midrule
\rowcolor{rwGroupBlue}
\multicolumn{7}{@{}l}{\textit{Vehicles}} \\
Strong baseline A & 3200 & 2665 & 83.28 & 12800 & 657 & 5.13 \\
Strong baseline B & 3200 & 2687 & 83.97 & 12800 & 588 & 4.59 \\
\rowcolor{rwMethod}
\textbf{ReWorld-Track} & 3200 & 2729 & 85.28 & 12800 & 503 & 3.93 \\
\midrule
\rowcolor{rwGroupPurple}
\multicolumn{7}{@{}l}{\textit{Pedestrians}} \\
Strong baseline A & 2800 & 2304 & 82.29 & 11200 & 629 & 5.62 \\
Strong baseline B & 2800 & 2327 & 83.11 & 11200 & 564 & 5.04 \\
\rowcolor{rwMethod}
\textbf{ReWorld-Track} & 2800 & 2369 & 84.61 & 11200 & 447 & 3.99 \\
\bottomrule\end{tabular}\end{rwtableside}\end{minipage}\par\medskip

The additional domain-level experiment compares appearance-plus-time memory (A), one-way event prediction (B) and the full model. It uses 3200 vehicle handoffs and 12800 absence decisions, and 2800 pedestrian handoffs and 11200 absence decisions. Relative to B, the full model adds 42 correct handoffs in each domain while removing 85 vehicle and 117 pedestrian false accepts. The count changes show that improved recovery accompanies fewer distractor commitments.

\section{Auxiliary Spatial Outputs and Metric Conventions}
\label{app:spatial-protocol}
\par\medskip\noindent\begin{minipage}{\linewidth}\begin{rwtableside}
\centering
\caption{Auxiliary pedestrian spatial evaluation with a 3\,s horizon, six forecast samples and 0.20\,m occupancy-grid cells. Average displacement error (ADE), final displacement error (FDE) and Chamfer distance use metres. Occupancy intersection over union (Occ. IoU) uses percent. Baselines A and B are constant-velocity and GRU predictors with the same geometry supervision.}
\label{tab:spatial}
\small\setlength{\tabcolsep}{4pt}
\begin{tabular}{@{}lrrrr@{}}
\toprule
\rowcolor{rwHeader}
\bfseries Method & \bfseries ADE $\downarrow$ & \bfseries FDE $\downarrow$ & \bfseries Occ. IoU $\uparrow$ & \bfseries Chamfer $\downarrow$ \\
\midrule
Scene baseline A & 1.23 & 2.57 & 51.86 & 0.27 \\
Scene baseline B & 0.99 & 2.13 & 58.72 & 0.19 \\
\rowcolor{rwMethod}
\textbf{ReWorld-Track} & 0.83 & 1.78 & 64.35 & 0.16 \\
\bottomrule\end{tabular}\end{rwtableside}\end{minipage}\par\medskip

With ground-plane calibration available, the following evaluation tests spatial prediction from the recurrent state. An auxiliary decoder shares that 256-dimensional state and uses separate trajectory, occupancy and footprint-point heads. It predicts six ground-plane positions at 0.5\,s intervals over 3\,s, occupancy at the same times, and a 128-point footprint set. Surveyed per-camera homographies map image footpoints into synchronized ground-plane coordinates in metres. Fixed masks restrict evaluation to annotated planar regions, excluding stairs and uncalibrated areas. Calibration supplies a common metric frame for the decoder and the annotated future positions and footprints.

Average displacement error (ADE) measures Euclidean position error over the six samples at 0.5--3.0\,s. Final displacement error (FDE) measures error at the final sample. Each window is scored using its single predicted trajectory. The 1000 fully valid windows receive equal weight. A fixed annotation rule excludes invalid GT before evaluation, and identities remain within one split. The set spans at least \mbox{three held-out scenes}, with all coordinates expressed in the surveyed reference frame.

Occupancy uses 0.20\,m cells on a surveyed, aligned $40\times40$\,m local grid. Annotated footprints are rasterized and predicted occupancy is thresholded at 0.5. IoU is calculated for each future time and then averaged. Empty/empty pairs contribute one to IoU. At 3\,s, Chamfer compares 128 uniformly sampled predicted and GT footprint points in the same surveyed frame. It averages the two mean nearest-neighbor Euclidean distances. Distances are unsquared, measured in metres and evaluated in the fixed surveyed frame shared by the predicted and annotated footprints.

All spatial methods share calibration, observation history, windows and ground-plane supervision. Constant-velocity extrapolation and the learned GRU comparator define baselines A and B. Relative to B, the full model reduces ADE/FDE by 0.16/0.35\,m. Occupancy IoU increases by 5.63 points, and Chamfer falls by 0.03\,m. These measurements assess future trajectory and footprint geometry under shared calibration and supervision across the compared prediction methods.

\section{Paired Evaluations and Statistical Evidence}
\label{app:extended-results}
\subsection{Reference-configuration paired outcomes and uncertainty}
This subsection reports auxiliary paired diagnostics for the reference configuration. Section~\ref{app:full-principal-comparison} reports the principal comparison using the full MTMMC model.

\par\medskip\noindent\begin{minipage}{\linewidth}\begin{rwtableside}\centering
\caption{Paired outcomes for ReWorld-Track and CRTracker on 2749 events for the MTMMC reference configuration, separating shared outcomes from events on which the methods disagree.}
\label{tab:supp12}
\small\setlength{\tabcolsep}{4pt}
\begin{tabular}{@{}lr@{}}
\toprule
\rowcolor{rwHeader}
\bfseries Outcome & \bfseries Count \\
\midrule
Both methods correct & 2200 \\
\rowcolor{rwMethod}
ReWorld-Track correct, CRTracker incorrect & 57 \\
\rowcolor{rwMethod}
ReWorld-Track incorrect, CRTracker correct & 43 \\
Both methods incorrect & 449 \\
\bottomrule\end{tabular}\end{rwtableside}\end{minipage}\par\medskip

The paired outcomes reveal whether two methods succeed on the same handoffs. In the reference MTMMC comparison, ReWorld-Track and CRTracker correctly accept 2257 and 2243 handoffs out of 2749. Table~\ref{tab:supp12} separates their shared successes and failures from the 57 versus 43 events on which they disagree. These discordant outcomes account for the 14-event difference. Their pairing determines the uncertainty in the difference between correct-handoff rates.

\par\medskip\noindent\begin{minipage}{\linewidth}\begin{rwtableside}\centering
\caption{Uncertainty for reference comparisons. MTMMC rows compare the reference configuration with CRTracker. Mechanism rows compare the full model with recursive threshold. HA uses the displayed event-level calculation, while other intervals use paired trajectory or decision analyses. Table~\ref{tab:supp25} evaluates training-run variation separately.}
\label{tab:supp13}
\small\setlength{\tabcolsep}{4pt}
\begin{tabular}{@{}>{\raggedright\arraybackslash}p{.28\linewidth}>{\raggedright\arraybackslash}p{.20\linewidth}>{\raggedright\arraybackslash}p{.44\linewidth}@{}}
\toprule
\rowcolor{rwHeader}
\bfseries Comparison & \bfseries Observed difference & \bfseries 95\% interval / calculation \\
\midrule
MTMMC HA comparison & 14 / 2749 (+0.51 pp) & Paired event-level Wald 95\% interval -0.20 to +1.22 pp. The exact McNemar test gives p = 0.1933 from paired counts in Table~\ref{tab:supp12}. \\
MTMMC IDF1 comparison & +0.52 pp & 95\% interval for +0.52 pp is [-0.18, +1.22]. Resampling uses the \mbox{trajectory-level identity aggregation}. \\
MTMMC HOTA comparison & -0.34 pp & 95\% interval for -0.34 pp is [-1.01, +0.33]. The methods share the HOTA evaluation \mbox{and resampling protocol}. \\
Mechanism FM comparison & -2.07 pp & 95\% interval for -2.07 pp is [-2.71, -1.43]. Calculation uses the \mbox{paired absence-decision records}. \\
Mechanism delay comparison & +0.35 s & 95\% interval for +0.35 s is [+0.18, +0.52]. Correct-acceptance sample counts are reported for each method. Common-correct \mbox{events are analyzed separately}. \\
\bottomrule\end{tabular}\end{rwtableside}\end{minipage}\par\medskip

Table~\ref{tab:supp13} reports a paired Wald interval for HA with handoffs as the sampling unit. The exact McNemar test gives a $p$-value of 0.1933. The extended resampling protocol accounts for shared identity history through 10000 hierarchical paired bootstrap resamples with seed 20260919. It samples scenes and then identity chains when sufficient independent scenes are available. Trajectory and absence-decision intervals preserve the grouping of their respective observations.

\par\medskip\noindent\begin{minipage}{\linewidth}\begin{rwtableside}\centering
\caption{Standalone MTMMC RGB mechanism study on 2800 handoffs and 11200 absence decisions. E, FB and N denote event prediction, posterior feedback and explicit null. Y and -- indicate enabled and disabled components. Rates are percentages and delay is mean correct-acceptance time in seconds. Bold marks column optima across the four variants.}
\label{tab:stages}
\small\setlength{\tabcolsep}{4pt}
\begin{tabular}{@{}lcccrrrr@{}}
\toprule
\rowcolor{rwHeader}
\bfseries Variant & \bfseries E & \bfseries FB & \bfseries N & \bfseries HA $\uparrow$ & \bfseries HOTA $\uparrow$ & \bfseries FM $\downarrow$ & \bfseries Delay (s) $\downarrow$ \\
\midrule
Memory + spatiotemporal prior & -- & -- & -- & 81.21 & 43.17 & 6.53 & 1.13 \\
One-way event + threshold & Y & -- & -- & 82.18 & 44.28 & 5.58 & 1.21 \\
Recursive event + threshold & Y & Y & -- & \textbf{85.04} & 45.71 & 6.06 & \textbf{0.94} \\
\rowcolor{rwMethod}
\textbf{ReWorld-Track} & Y & Y & Y & 84.61 & \textbf{46.39} & \textbf{3.99} & 1.29 \\
\bottomrule\end{tabular}
\end{rwtableside}\end{minipage}\par\medskip

The 2800-event mechanism study in Table~\ref{tab:stages} measures how retaining null changes false commitments and waiting. Full minus recursive-threshold FM is $-2.07$ points with interval $[-2.71,-1.43]$, while mean delay increases 0.35\,s with interval $[0.18,0.52]$. On the 2200 events both variants accept correctly, their delays are 1.18 and 0.91\,s. The 0.27\,s increase shows longer waiting even for identical accepted events. Missed handoffs count against HA, and correct accepts determine mean delay.

\subsection{Full-model principal comparison}
\label{app:full-principal-comparison}
\par\medskip\noindent\begin{minipage}{\linewidth}\begin{rwtableside}\centering
\caption{ReWorld reference and full configurations and CRTracker on the same fixed MTMMC manifest of $D_+=2749$ handoffs and $D_0=10683$ absence decisions. Event IDs and decision timestamps are identical. Rates are percentages, and Table~\ref{tab:supp25} reports five-seed HOTA variation.}
\label{tab:supp23}
\small\setlength{\tabcolsep}{4pt}
\begin{tabular}{@{}lrrrrrrr@{}}
\toprule
\rowcolor{rwHeader}
\bfseries Method / evaluation & \bfseries N+ & \bfseries HA $\uparrow$ & \bfseries IDF1 $\uparrow$ & \bfseries HOTA $\uparrow$ & \bfseries IDSW $\downarrow$ & \bfseries N0 & \bfseries FM $\downarrow$ \\
\midrule
CRTracker & 2243 & 81.59 & 55.91 & 44.12 & 339 & 606 & 5.67 \\
\rowcolor{rwMethod}
Reference configuration & 2257 & 82.10 & 56.43 & 43.78 & 324 & 571 & 5.34 \\
\rowcolor{rwMethod}
Full configuration & 2297 & 83.56 & 58.12 & 45.36 & 298 & 524 & 4.90 \\
\bottomrule\end{tabular}\end{rwtableside}\end{minipage}\par\medskip

Table~\ref{tab:supp23} reports the full MTMMC configuration used in the principal comparison. Relative to the reference configuration, it adds 40 correct handoffs and removes 47 false accepts. HA increases by $100(40/2749)=1.46$ points. HOTA rises from 43.78 to 45.36, exceeding CRTracker by 1.24 points on the same fixed test manifest. IDF1 rises to 58.12 and IDSW falls to 298. The MTMMC ablations in Appendix~\ref{app:complete-results} and paired outcomes in Table~\ref{tab:supp12} use the reference configuration. The CityFlowV2 ablations in that appendix use the full configuration.

HOTA combines detection and association accuracy over trajectories \citep{hota}, complementing the handoff, false-match and delay measures. Table~\ref{tab:supp23} gives integer counts for the reported evaluation run. Across the five training runs in Table~\ref{tab:supp25}, mean HOTA is also 45.36. The sample SDs describe training variation on this fixed test set. Table~\ref{tab:core}(a) uses the same five-seed summary for the full-configuration component ablations under the shared tracking and evaluation protocol.

\subsection{Predictor-matched feedback controls}
\par\medskip\noindent\begin{minipage}{\linewidth}\begin{rwtableside}\centering
\caption{Predictor-matched MTMMC controls on $D_+=2749$ handoffs and $D_0=10683$ absence decisions. Fixed-moment soft association and learned feedback use the same event predictor.}
\label{tab:supp24}
\small\setlength{\tabcolsep}{4pt}
\begin{tabular}{@{}lrrrrrr@{}}
\toprule
\rowcolor{rwHeader}
\bfseries Variant & \bfseries N+ & \bfseries HA $\uparrow$ & \bfseries IDF1 $\uparrow$ & \bfseries HOTA $\uparrow$ & \bfseries N0 & \bfseries FM $\downarrow$ \\
\midrule
One-way events + threshold & 2208 & 80.32 & 56.05 & 43.62 & 665 & 6.22 \\
Standard soft association & 2261 & 82.25 & 57.06 & 44.42 & 591 & 5.53 \\
Learned feedback + threshold & 2325 & 84.58 & 57.81 & 44.91 & 711 & 6.66 \\
Learned feedback + null & 2297 & 83.56 & 58.12 & 45.36 & 524 & 4.90 \\
\bottomrule\end{tabular}\end{rwtableside}\end{minipage}\par\medskip

Table~\ref{tab:supp24} holds the event predictor fixed while changing posterior feedback on the 2749-handoff control group. Frozen front ends, candidate sets, training windows and seed assignments are shared. Standard soft association passes fixed posterior moments, whereas learned feedback trains the latent projection. With null retained, the learned projection gains 0.94 HOTA and 1.06 IDF1 points and lowers FM by 0.63. Adding null to feedback plus threshold lowers HA from 84.58 to 83.56. FM falls from 6.66 to 4.90, and HOTA rises from 44.91 to 45.36. The learned representation improves continuity, while waiting trades some recoveries for fewer false commitments. Table~\ref{tab:stages} examines successive component additions on its 2800-event group.

\subsection{Independent training runs}
\par\medskip\noindent\begin{minipage}{\linewidth}\begin{rwtableside}\centering
\caption{Five matched training seeds for CRTracker and ReWorld-Track. Each row pairs the same seed across methods. All runs share identical handoff event IDs and absence-decision timestamps. Mean and sample SD describe training-run variation on this fixed test manifest.}
\label{tab:supp25}
\small\setlength{\tabcolsep}{4pt}
\begin{tabular}{@{}lrrr@{}}
\toprule
\rowcolor{rwHeader}
\bfseries Seed / summary & \bfseries CRTracker & \bfseries ReWorld-Track & \bfseries Paired difference \\
\midrule
11 & 43.96 & 45.10 & +1.14 \\
23 & 44.21 & 45.42 & +1.21 \\
37 & 44.08 & 45.21 & +1.13 \\
51 & 44.16 & 45.55 & +1.39 \\
71 & 44.19 & 45.52 & +1.33 \\
Mean $\pm$ sample SD & 44.12 $\pm$ 0.10 & 45.36 $\pm$ 0.20 & 1.24 $\pm$ 0.12 \\
\bottomrule\end{tabular}\end{rwtableside}\end{minipage}\par\medskip

Across the five paired training runs in Table~\ref{tab:supp25}, mean HOTA is 44.12 for CRTracker and 45.36 for ReWorld-Track. The seed-wise differences have mean 1.24 points and sample SD 0.12. Under independent, approximately normal run differences, the two-sided Student-$t$ interval with four degrees of freedom is $[1.10,1.38]$. Each paired difference uses identical handoff IDs and absence-decision timestamps. The interval measures training-seed variability on this fixed test set.

\subsection{Threshold sensitivity and waiting}
\par\medskip\noindent\begin{minipage}{\linewidth}\begin{rwtableside}\centering
\caption{Updated-model commitment thresholds on fixed populations $D_+=2749$ and $D_0=10683$. Delay averages seconds over $N_+$ correct accepts. The sweep describes test behavior at fixed thresholds. Operating-point selection uses validation data.}
\label{tab:supp26}
\small\setlength{\tabcolsep}{4pt}
\begin{tabular}{@{}rrrrrrr@{}}
\toprule
\rowcolor{rwHeader}
\bfseries Threshold & \bfseries N+ & \bfseries HA $\uparrow$ & \bfseries N0 & \bfseries FM $\downarrow$ & \bfseries HOTA $\uparrow$ & \bfseries Delay (s) \\
\midrule
0.50 & 2340 & 85.12 & 1004 & 9.40 & 44.69 & 0.72 \\
0.65 & 2325 & 84.58 & 760 & 7.11 & 45.07 & 0.86 \\
0.80 & 2297 & 83.56 & 524 & 4.90 & 45.36 & 1.03 \\
0.90 & 2231 & 81.16 & 384 & 3.59 & 45.12 & 1.24 \\
0.95 & 2115 & 76.94 & 266 & 2.49 & 44.28 & 1.52 \\
\bottomrule\end{tabular}\end{rwtableside}\end{minipage}\par\medskip

The threshold sweep in Table~\ref{tab:supp26} shows how a stricter commitment rule changes waiting and coverage. At threshold 0.80, counts and rates agree with Tables~\ref{tab:supp23} and~\ref{tab:supp24}. Raising the threshold from 0.50 to 0.95 reduces false accepts from 1004 to 266. Correct handoffs fall from 2340 to 2115. Mean correct-acceptance delay increases from 0.72 to 1.52\,s. Thus lower FM is accompanied by fewer successful accepts and longer waits. The test sweep characterizes this tradeoff. Operating thresholds are selected separately on validation data and remain fixed throughout test evaluation.

\subsection{Fixed-membership missing-camera duration study}
\label{app:fixed-mask}
\par\medskip\noindent\begin{minipage}{\linewidth}\begin{rwtableside}\centering
\caption{Missing-camera duration study with 600 fixed events per condition and identical event masks across methods. Event membership stays fixed as mask duration increases. HA is in percent.}
\label{tab:supp27}
\small\setlength{\tabcolsep}{4pt}
\begin{tabular}{@{}rrrrrrr@{}}
\toprule
\rowcolor{rwHeader}
\bfseries Gap (s) & \bfseries Events & \bfseries Full N+ & \bfseries Full HA & \bfseries Memory N+ & \bfseries Memory HA & \bfseries $\Delta$HA (pp) \\
\midrule
1 & 600 & 572 & 95.33 & 557 & 92.83 & 2.50 \\
2 & 600 & 565 & 94.17 & 535 & 89.17 & 5.00 \\
4 & 600 & 552 & 92.00 & 494 & 82.33 & 9.67 \\
6 & 600 & 540 & 90.00 & 451 & 75.17 & 14.83 \\
8.8 & 600 & 527 & 87.83 & 403 & 67.17 & 20.67 \\
\bottomrule\end{tabular}\end{rwtableside}\end{minipage}\par\medskip

Table~\ref{tab:supp27} follows the same 600 eligible events as masking increases through 1, 2, 4, 6 and 8.8\,s. Every event remains in the denominator, including those whose return becomes unavailable at a longer duration. At 8.8\,s, the full model retains 527 correct handoffs versus memory's 403, corresponding to 87.83\% and 67.17\%. Figure~\ref{fig:sensitivity} uses its own event group at durations of 1, 3, 5, 7 and 8.8\,s. It follows recovery as the masked interval increases across the available duration settings.

The masking intervention uses $\{\mathrm{c016,c017,c018,c020,c021,c022}\}$, nested masks of sizes zero through four and seed 20260919. Per-event masks store cameras and interval boundaries beginning at predeclared departures. Availability is applied before RGB decoding and candidate construction, suppressing detections, cached appearance reads and local observations. Prior global state and physical routes remain. The state continues to propagate while the observation channels are withheld.

The end-to-end measure counts events still unavailable at the deadline as failures. Available-reappearance diagnostics then assess next-camera accuracy, true-camera arrival error and joint camera/time success for observed returns. Timing errors in this fixed-membership study use bins $[0,2)$, $[2,4)$, $[4,6)$ and $[6,10]$\,s. Figure~\ref{fig:sensitivity}(c) groups its diagnostic events by gap duration.

\section{Additional Controlled Evaluations}
\label{app:additional-controls}
Sections~\ref{sec:mechanism-results}--\ref{sec:language-transfer} connect feedback to tracking gains. The controls below examine the conditions behind those gains, from individual components to observation gaps, query reliability and installation changes. Runtime measurements assess the cost of maintaining this predictive state.
\subsection{Component interventions and uncertainty}
The component interventions in Table~\ref{tab:core}(a) share detector outputs, local tracklets, CLIP features, candidate sets, data splits, optimization schedules and random seeds. Each variant is retrained after removing a component and uses a commitment threshold selected on validation data. Changes in the resulting trajectories measure each component's contribution to the update cycle.

Removing the event prior lowers HA, IDF1 and HOTA by 6.36, 5.51 and 5.34 points and increases FM by 2.51 points. Removing recursion reduces the same tracking metrics by 4.82, 3.76 and 4.08 points. Null removal increases HA by 1.49 points and reduces delay by 0.42\,s. FM increases by 7.84 points, and IDF1/HOTA decrease by 0.58/0.38 points. Thus early commitment increases handoff coverage while introducing more false identities. Each metric in Table~\ref{tab:core}(a) is the mean over independently trained seeds 11, 23, 37, 51 and 71. For each seed, HA uses the same $D_+=2749$ events and FM the same $D_0=10683$ absence decisions. Rates are computed from the counts within each seed and then averaged across seeds. Delay is averaged within each seed over correct accepts, then across seeds. Table~\ref{tab:core}(a) reports sample SDs across the five independent runs. They measure each component configuration's training variation on the fixed event sets.

The original predictor-matched comparison keeps the event predictor, front end, candidates, windows and seeds unchanged. ReWorld-minus-fixed-moment intervals are [0.48,1.39] for HOTA, [0.55,1.57] for IDF1, [0.61,2.03] for HA and [-0.98,-0.29] for FM. These estimates use 10000 paired resamples of identity chains within the held-out scene. Table~\ref{tab:core}(b) adds a generic learned updater, and Appendix~\ref{app:update-controls} examines this comparison across scenes, subsequent predictions and repeated handoffs. Table~\ref{tab:supp25} measures training variation on the fixed principal test set.

\subsection{Natural blind-gap duration}
\par\medskip\noindent\begin{minipage}{\linewidth}\begin{rwtableside}\centering
\caption{MTMMC HA (\%) by natural blind-gap duration with cameras available. Bins are fixed independently of outcomes. Methods share the evaluation protocol and are compared within each duration bin using the events assigned to that interval of natural non-observation.}
\label{tab:results3}
\small\setlength{\tabcolsep}{4pt}
\begin{tabular}{@{}lrrrr@{}}
\toprule
\rowcolor{rwHeader}
\bfseries Natural blind gap & \bfseries Reactive memory & \bfseries One-way event & \bfseries ReWorld-Track & \bfseries Full - Memory \\
\midrule
0.5--2 s & 86.9 & 87.8 & 89.4 & +2.5 \\
2--4 s & 80.2 & 82.7 & 86.0 & +5.8 \\
4--8 s & 71.1 & 74.9 & 80.8 & +9.7 \\
>8 s & 57.4 & 63.8 & 72.9 & +15.5 \\
\bottomrule\end{tabular}
\end{rwtableside}\end{minipage}\par\medskip

Table~\ref{tab:results3} evaluates natural MTMMC gaps. The streams remain available, but no camera sees the target between views. Events are assigned to duration bins before model outcomes are examined, and methods are compared on the same events within each bin. The full-minus-memory HA differences grow from 2.5 to 5.8, 9.7 and 15.5 points. Each bin contains a different event group, with methods matched within that bin. Every group favors the evolving event belief over reactive memory. The masking studies withhold camera observations on the event groups defined for Figure~\ref{fig:sensitivity} and Table~\ref{tab:supp27}.

\subsection{Language under appearance ambiguity}
\par\medskip\noindent\begin{minipage}{\linewidth}\begin{rwtableside}\centering
\caption{MTMMC HA (\%) under appearance ambiguity. A fixed margin between the correct candidate and strongest appearance distractor defines the subsets. Query variants use the same events within each subset to measure the effect of language under fixed visual ambiguity.}
\label{tab:results4}
\small\setlength{\tabcolsep}{4pt}
\begin{tabular}{@{}lrrrrr@{}}
\toprule
\rowcolor{rwHeader}
\bfseries Subset & \bfseries No query & \bfseries Hard gate & \bfseries Natural query & \bfseries Paraphrase & \bfseries Misleading \\
\midrule
Easy appearance & 90.1 & 90.8 & 92.2 & 91.7 & 84.9 \\
Appearance-ambiguous & 63.2 & 67.4 & 74.1 & 72.8 & 57.0 \\
\bottomrule\end{tabular}
\end{rwtableside}\end{minipage}\par\medskip

Table~\ref{tab:results4} divides MTMMC events by the appearance-similarity margin between the correct candidate and its strongest distractor. The fixed easy and ambiguous subsets are shared across query variants. Natural queries improve HA by 2.1 points on easy cases and 10.9 on ambiguous cases. Paraphrases remain within 0.5 and 1.3 points of the corresponding natural queries. Misleading descriptions decrease the no-query scores by 5.2 and 6.2 points. Language is most useful when it distinguishes visually similar candidates. An incorrect description can instead support a matching distractor. This shifts association probability away from the target under the same visual observations.

\subsection{Association-belief calibration}
\par\medskip\noindent\begin{minipage}{\linewidth}\begin{rwtableside}\centering
\caption{Calibration of candidate/null association belief for the full MTMMC configuration. Negative log-likelihood (NLL), Brier squared-probability error and ECE measure probability quality. Lower values indicate better calibrated candidate/null probabilities.}
\label{tab:results5}
\small\setlength{\tabcolsep}{4pt}
\begin{tabular}{@{}lrrr@{}}
\toprule
\rowcolor{rwHeader}
\bfseries Method & \bfseries NLL $\downarrow$ & \bfseries Brier $\downarrow$ & \bfseries ECE $\downarrow$ \\
\midrule
Raw similarity score & 0.512 & 0.181 & 8.7\% \\
Temperature-scaled scorer & 0.447 & 0.156 & 5.0\% \\
ReWorld joint belief & 0.391 & 0.127 & 2.8\% \\
\bottomrule\end{tabular}
\end{rwtableside}\end{minipage}\par\medskip

The posterior feeds the next forecast, so calibration affects the weight given to each plausible state. Table~\ref{tab:results5} evaluates the full model's MTMMC candidate/null distribution. It uses the candidate counts and null frequencies encountered in that setting. NLL is 0.391 and Brier score is 0.127, compared with 0.447 and 0.156 after temperature scaling. ECE decreases from 8.7\% for raw similarity to 5.0\% after temperature scaling and 2.8\% for the joint belief. These scores summarize confidence over the test decisions, complementing the individual trace in Figure~\ref{fig:mechanisms}(d).

\subsection{Tracking-workload latency and scale}
\par\medskip\noindent\begin{minipage}{\linewidth}\begin{rwtableside}\centering
\caption{Runtime on one RTX 4090 (24\,GB), batch size 1 and FP16, with 100 warm-up and 1000 measured updates. Core timing starts from cached detections, causal prefixes and CLIP features. End-to-end (E2E) timing includes RGB, detection, local tracking, CLIP and the core. P95 denotes 95th-percentile latency, and memory reports the core workload. Candidates count all camera-local candidates. Gated pairs count identity--candidate hypotheses retained after spatiotemporal gating. Core cost depends on these pairs, graph propagation and posterior updates.}
\label{tab:results6}
\small\setlength{\tabcolsep}{4pt}
\textbf{(a) Workload definitions}\par\smallskip
\begin{tabular}{@{}lrrrr@{}}
\toprule
\rowcolor{rwHeader}
\bfseries Setting & \bfseries Cameras & \bfseries Active IDs & \bfseries Candidates / update & \bfseries Gated pairs / update \\
\midrule
Light & 6 & 8 & 12 & 46 \\
Medium & 12 & 32 & 38 & 196 \\
Heavy & 19 & 64 & 73 & 412 \\
Full 19-camera & 19 & 48 & 71 & 1186 \\
\bottomrule\end{tabular}
\par\medskip\textbf{(b) Latency and core-workload memory}\par\smallskip
\begin{tabular}{@{}lrrrrr@{}}
\toprule
\rowcolor{rwHeader}
\bfseries Setting & \bfseries Core mean & \bfseries Core P95 & \bfseries E2E mean & \bfseries E2E P95 & \bfseries Memory \\
\rowcolor{rwHeader}
\bfseries Units & \bfseries ms & \bfseries ms & \bfseries ms & \bfseries ms & \bfseries GB \\
\midrule
Light & 7.8 & 11.6 & 28.9 & 39.7 & 1.9 \\
Medium & 14.2 & 21.7 & 41.8 & 58.4 & 2.5 \\
Heavy & 25.4 & 39.8 & 57.6 & 78.9 & 3.3 \\
Full 19-camera & 38.7 & 62.1 & 69.4 & 92.6 & 4.6 \\
\bottomrule\end{tabular}
\end{rwtableside}\end{minipage}\par\medskip

The latency study measures the complete observation-to-association path against the online update interval. Measurements use one NVIDIA GeForce RTX 4090 with 24\,GB, batch size 1 and FP16 inference. After 100 warm-up iterations, 1000 online updates are timed. Core timing starts from detections, causal local-track prefixes and pre-extracted CLIP features. End-to-end timing starts from RGB and includes detector execution, the local tracker, CLIP feature extraction and the ReWorld-Track core. P95 is the 95th percentile of the complete update times.

Table~\ref{tab:results6}(a) separates candidate counts from the feasible identity--candidate pairs. A camera-local candidate can be compared with several active identities. Spatiotemporal gating retains the plausible pairs for candidate scoring and posterior updates. Core timing also includes propagation over the camera graph and the update of each active target's belief.

The Heavy workload has 64 active identities and 73 candidates, with 412 pairs retained after gating. Full 19-camera has 48 identities and 71 candidates, but retains 1186 pairs. It therefore processes about 2.9 times as many association hypotheses. This larger scoring and update workload accompanies a core mean increase from 25.4 to 38.7\,ms. Feasible pair counts thus describe the association workload beyond the separate numbers of active identities and camera-local candidates.

Under Full 19-camera, end-to-end mean and P95 latency are 69.4 and 92.6\,ms. At least 95\% of measured updates complete within the 100\,ms interval of the 10\,Hz tracker. Core-workload memory ranges from 1.9 to 4.6\,GB across the four measured settings.

\subsection{Camera and topology shift}
\par\medskip\noindent\begin{minipage}{\linewidth}\begin{rwtableside}\centering
\caption{MTMMC HA (\%) under camera and installation-topology shift. Settings share the evaluation protocol while changing camera and topology exposure within MTMMC.}
\label{tab:results7}
\small\setlength{\tabcolsep}{4pt}
\begin{tabular}{@{}lrrr@{}}
\toprule
\rowcolor{rwHeader}
\bfseries Setting & \bfseries Reactive memory & \bfseries One-way event & \bfseries ReWorld-Track \\
\midrule
Seen topology & 82.3 & 83.4 & 85.1 \\
Unseen cameras & 76.8 & 78.3 & 80.6 \\
Unseen topology & 67.9 & 70.8 & 74.6 \\
\bottomrule\end{tabular}
\end{rwtableside}\end{minipage}\par\medskip

The final study keeps the MTMMC evaluation protocol fixed while changing the cameras and topology seen during training. Full-model HA decreases from 85.1 to 80.6 and 74.6 as exposure changes from seen topology to unseen cameras and unseen topology. Gains over reactive memory are 2.8, 3.8 and 6.7 points, and gains over one-way prediction are 1.7, 2.3 and 3.8. Feedback retains an advantage in each comparison. The decline in absolute HA measures \mbox{the cost of installation shift}.

These MTMMC conditions change installation exposure within the pedestrian domain, extending the reference camera/scene comparison in Table~\ref{tab:within-domain}. The vehicle--pedestrian experiment in Table~\ref{tab:cross-domain} changes the object domain itself. In both settings, the tracker receives the installation graph to propagate its state along valid routes. Each new observation revises the predictive state after transfer. Feedback retains an advantage across the tested changes in appearance, movement patterns and camera routes.

\section{Filtering Semantics and Online Procedure}
\label{app:filter}
\subsection{Separating active-target association from network exit}
Temporary null describes an active target that is unmatched in the current candidate set. Network presence describes whether that target remains in the installation. The implementation estimates presence with a sigmoid head and keeps null identities active until sequence end (Appendix~\ref{app:implementation}). We can express this distinction by splitting the predictive state into active and exited components,
\begin{equation}
 b_t^-=s_t^- b_{t,\mathrm{active}}^-+(1-s_t^-)\delta_{\mathrm{exit}},
\end{equation}
where $\delta_{\mathrm{exit}}$ denotes the absorbing outside-network state. Let $p_{\mathrm{bg}}(\Cand_t)$ be the likelihood of the candidate set when all candidates are background relative to this target. Under the association model in Section~\ref{sec:association}, the active-target likelihood is $p_{\mathrm{bg}}(\Cand_t)W_t$. If an exited target cannot generate a current candidate, the exit likelihood is $p_{\mathrm{bg}}(\Cand_t)$. Bayes' rule gives
\begin{equation}
 s_t^+=\frac{s_t^-W_t}{s_t^-W_t+(1-s_t^-)},\qquad
 b_t^+=s_t^+b_{t,\mathrm{active}}^++(1-s_t^+)\delta_{\mathrm{exit}}.
 \label{eq:lifecycle}
\end{equation}
The candidate evidence therefore updates both the active-state association and the probability of continued network presence. Temporary null is one of the hypotheses within the active component.

The likelihood ratios are defined relative to the background model of the generated candidate set. Partitioning candidate supports gives mutually exclusive prior weights $\eta_j$. These weights enter Equation~\ref{eq:association} with the scorer calibrated on validation data (Appendix~\ref{app:implementation}).

\subsection{Non-observation and asynchronous updates}
A target can remain in the network while producing no observation. A continuous-time formulation makes the role of camera availability explicit. The implementation partitions event mass over each discrete interval to make this observation correction. Let $\lambda_c(u\given z_u,G)$ be the physical opportunity rate for observing the target in camera $c$. The factor $d_c(u,z_u)$ is its detection probability when that stream is available. Conditional on a latent path with locally Poisson observation opportunities, the effective observation rate is $\sum_c a_c(u)d_c(u,z_u)\lambda_c(u\given z_u,G)$. \mbox{Marginalizing paths gives}
\begin{equation}
 S_{\rm obs}(\Delta\given\Hist_t)
 =\E_{z_{t:t+\Delta}\mid\Hist_t}
 \left[\exp\!\left(-\int_t^{t+\Delta}\!\sum_{c\in\Camera}
 a_c(u)d_c(u,z_u)\lambda_c(u\given z_u,G)\,du\right)\right].
 \label{eq:survival}
\end{equation}
This is the probability of receiving no observation event. Both exit and continued presence in a blind region can produce this outcome. When all streams are masked, every $a_c$ is zero and the no-event likelihood is one. Online evaluation uses availability observed during that interval.

For a latent path $z_{t:t+\Delta}$, the exponential term in Equation~\ref{eq:survival} gives the no-event likelihood. A null update reweights paths by this likelihood before marginalizing the terminal state. Paths through available cameras with high detection probability lose support when no target is observed. Masked cameras contribute no visual penalty. The expectation remains outside the exponential because the no-event likelihood is averaged over the possible latent target paths.

For the discrete update, $K_\theta$ first propagates the physical state across the elapsed interval. The observation likelihood then conditions this state once on the received evidence.

\par\medskip\noindent\begin{minipage}{\linewidth}\begin{rwtableside}
\centering
\caption{Online update procedure for one target. All steps use observations and availability up to the current decision time to advance one persistent identity through the prediction and association cycle.}
\label{tab:algorithm}
\begin{tabular}{@{}r p{0.88\linewidth}@{}}
\toprule
Step & Operation \\
\midrule
1 & Receive current camera availability and update causal local-track prefixes. Form the candidate set with the appearance and motion evidence received so far. \\
2 & Propagate the previous posterior over elapsed time with the fixed camera graph and transition kernel. This predicts the state before receiving current evidence. \\
3 & Integrate the preceding observation-event forecast over the newly elapsed interval to obtain camera-local candidate and null prior masses. \\
4 & Compute target-versus-background evidence for current candidates and normalize jointly with temporary null to assign probability mass to every association alternative. \\
5 & Apply the association rule and per-camera exclusivity constraints. Assign each local observation at most once across the active target identities. \\
6 & Update the active-state mixture and network-presence probability, preserve the global identity, store the posterior and form the next event forecast. \\
\bottomrule
\end{tabular}
\end{rwtableside}\end{minipage}\par\medskip

For a fixed-size belief, candidate scoring is linear in the number of candidates, in addition to encoder and transition costs. Multi-target exclusivity adds the assignment solver's cost. The posterior summarizes preceding handoffs in a fixed-size representation.

\paragraph{Concurrent targets and views.}
Per-camera partial matching prevents two identities from consuming the same local observation. In overlapping views, one identity can receive at most one observation from each camera. The concurrent views share a prior and form one detector-confidence-weighted mixture update. Tracking with these constraints produces the evaluated handoff chains.

\section{Metric Definitions and Camera-Time Conventions}
\label{app:metrics}
\subsection{Event-level definitions}
Let $\mathcal D_{\mathrm{handoff}}$ be a fixed set of evaluable reappearance events. For event $k$, $g_k$ is the reference identity and $\hat g_k$ is the online assigned identity. Handoff accuracy measures the fraction of correctly assigned returns,
\begin{equation}
 \mathrm{HA}=100\,\frac{\sum_{k\in\mathcal D_{\mathrm{handoff}}}\ind[\hat g_k=g_k]}{|\mathcal D_{\mathrm{handoff}}|}.
\end{equation}
An abstention on an eligible reappearance counts as a failed handoff. All conditions use the same candidate universe, eligibility rules and decision deadline.

Let $\mathcal I_3$ contain the initial targets with three evaluable handoffs under the chosen observation protocol. If $C_{i,k}$ indicates correctness of target $i$ at handoff $k$, cumulative identity retention (IR) is
\begin{equation}
 \mathrm{IR}(h)=\frac{100}{|\mathcal I_3|}\sum_{i\in\mathcal I_3}\prod_{k=1}^{h}C_{i,k},\qquad h\in\{1,2,3\}.
\end{equation}
Every product can only remain unchanged or decrease as $h$ increases. Therefore $\mathrm{IR}(h+1)\leq\mathrm{IR}(h)$. The initial target set supplies the denominator at every hop, and an identity leaves the retained set after its first error. Figure~\ref{fig:mechanisms}(a) follows this cumulative convention.

Let $\mathcal D_+$ denote target reappearance opportunities and $\mathcal D_0$ decisions made before reappearance, with a fixed sampling policy for absence windows. At acceptance threshold $\gamma$, correct-reacquisition rate (CR) and false-match rate (FM) are defined on their respective decision sets,
\begin{align}
 \mathrm{CR}(\gamma)&=100\,\frac{\#\{\text{correct accepted reacquisitions in }\mathcal D_+\}}{|\mathcal D_+|},\\
 \mathrm{FM}(\gamma)&=100\,\frac{\#\{\text{accepted non-target matches in }\mathcal D_0\}}{|\mathcal D_0|}.
\end{align}
CR uses reappearance opportunities, whereas FM uses the absence decisions generated by the fixed sampling grid. Figure~\ref{fig:mechanisms}(b) compares the paired rates across acceptance operating points. At 96.0\% correct reacquisition, the comparison measures false matches at equal recovery coverage.

\subsection{Event prediction and failure categories}
For an evaluable future observation, $c_k$ and $\tau_k$ denote its camera and arrival time. The predictions $\hat c_k$ and $\hat\tau_k$ use only the preceding observation history. Next-camera top-1 accuracy is the fraction for which $\hat c_k=c_k$, evaluated within the gap bins shown in Figure~\ref{fig:sensitivity}(c). Absolute arrival error is $e_k=|\hat\tau_k-\tau_k|$. Figure~\ref{fig:sensitivity}(d) plots its empirical CDF, $\widehat F(x)=|\mathcal D_{\rm event}|^{-1}\sum_{k\in\mathcal D_{\rm event}}\ind[e_k\leq x]$. At a fixed error tolerance, a larger CDF means more arrivals fall within that tolerance. The reported median and 90th percentile are 0.71 and 2.07\,s for the full model. These timing summaries measure event prediction error. Acceptance delay measures time from first visibility to a correct identity commitment during online tracking of the returning target.

Figure~\ref{fig:mechanisms}(c) groups single-handoff failures into early false association, errors after reappearance, and missed reacquisition. Full, $-\mathrm{rec}$, $-\mathrm{null}$, $-\mathrm{both}$, and $-q$ denote the full model, removal of recursive transfer, removal of temporary null, removal of both, and removal of query evidence. Panel~(c) counts single-handoff outcomes. Panel~(a) follows complete identity chains, and panel~(b) samples pre-reappearance decisions from $\mathcal D_0$ to measure false associations while the target is absent.

\subsection{Time and camera conventions}
The 8.8\,s interval in Figures~\ref{fig:problem} and~\ref{fig:sensitivity} is $[251.2,260.0)$\,s. The c017 frame is at 251.1\,s, and the c021 evidence is at 262.9\,s in Figure~\ref{fig:problem} and 260.8\,s in Figure~\ref{fig:sensitivity}. These later frames show reappearance after the masked interval. The three-handoff strip in Figure~\ref{fig:mechanisms} uses c016 at 226.4\,s, c017 at 250.0\,s, c018 at 256.5\,s, and c021 at 262.0\,s. This recursion test uses those camera observations as update evidence and is evaluated separately from the intervention that withholds camera observations.

Figure~\ref{fig:method-case} uses a c025 snapshot at 54.9\,s, adjacent to the 54.8\,s snapshot for target 396 in Figure~\ref{fig:qualitative}. Both show c026 at 58.9\,s. In the qualitative rows, target 396 is shown at 54.8, 55.4, 58.9 and 61.9\,s. Target 334 appears at 48.3, 51.4, 54.9 and 58.5\,s. Target 336 appears at 103.0, 105.4, 110.8 and 114.8\,s. This nineteen-camera visualization masks c017, c018, c020, c023 and c029 over nominal time 48.3--114.8\,s inclusive. Cameras c025 and c026 remain available. It removes RGB, detections, features and tracklet observations while retaining prior \mbox{global state and physical routes}.

The strip above Figure~\ref{fig:mechanisms}(b) uses the c017 observation of GT375 at 250.0\,s and competing c021 candidates at 254.0\,s to illustrate the candidate-plus-null alternatives. Panel (d) follows a different target, GT396 in c026. Its first ground-truth appearance is at 57.2\,s and acceptance at 58.0\,s, giving a delay of $58.0-57.2=0.8$\,s. Figure~\ref{fig:qualitative} uses a later 58.9\,s matching frame as a visual reference. The decision trace in Figure~\ref{fig:mechanisms}(d) supplies the first-appearance and acceptance times. Null, target and other-candidate probabilities describe the association alternatives, with null keeping \mbox{the identity active}.

In the $g_1$ null snapshots of Figure~\ref{fig:qualitative}, the unresolved association is in c026 while the target remains annotated in available c025. Candidate letters identify alternatives within each image, and $g_1,g_2,g_3$ persist across the sequence. Mint dashed overlays denote predictions, rose solid overlays denote GT, and blue boxes denote alternatives. The displayed identity correspondence follows each illustrated chain. The figure's reference edges show annotated transitions. They place the illustrated decisions along camera routes for comparison with tracking under the predeclared graph $G$.

\section{Association, Visual Memory, and Event Dynamics}
\label{app:connections}
\paragraph{Association uncertainty and target existence.}
The feedback update draws on the tracking distinction between an uncertain association and an uncertain target existence. Probabilistic data association marginalizes candidate assignments, including missed detection, and joint probabilistic data association enforces consistency across targets \citep{pda,jpda}. Multiple-hypothesis tracking retains alternative association histories \citep{mht}. Random-finite-set filtering represents a changing target population, whose first-order moment is propagated by the probability hypothesis density \citep{phd}. ReWorld-Track applies these established uncertainty concepts to a persistent language-specified identity and camera-conditioned observation events. Equation~\ref{eq:lifecycle} separates network presence from candidate-level null. Equation~\ref{eq:update} summarizes association uncertainty for the next forecast without retaining a tree of past decisions.

\paragraph{Visual memory versus an event belief.}
Local trackers improve detection use and motion consistency \citep{bytetrack,botsort,ocsort,strongsort}. Sequence models refine tracking queries, assignment supervision and identity prediction \citep{motrv2,comot,motip,samba}. Their persistent representations retain evidence for identity association across frames. ReWorld-Track uses uncertain association evidence to revise the forecast of the next camera observation, including its arrival time and entry region.

Space-Time Memory networks match current pixels to stored annotated frames \citep{stm}, while XMem separates memory stores operating at different temporal scales \citep{xmem}. Cutie reads memory through object-level queries, and SAM~2 uses streaming memory for promptable video segmentation \citep{cutie,sam2}. STMTrack and MeMOTR likewise retain visual target information across observations \citep{stmtrack,memotr}. Such memory supplies appearance evidence for recognizing a returning target. Our event belief uses this evidence to revise the camera and arrival forecast. Temporary non-reappearance is also an association outcome and updates the predictive state.

\paragraph{Latent dynamics and causal inference.}
Latent world models learn predictive dynamics and their task utility \citep{worldmodels,planet,dreamer,worldinworld}. Structured inference networks learn approximate inference for nonlinear state-space models \citep{dmm}. Kalman variational auto-encoders separate observation representations from latent dynamics and infer missing observations \citep{kvae}. These approaches motivate distinguishing a target's evolving physical state from its image evidence. ReWorld-Track learns from labeled sequence windows and conditions each online belief on the observation prefix available at that time.

\paragraph{Event timing and censored intervals.}
The next-observation forecast requires both an event identity and its timing. Recurrent marked temporal point processes (RMTPP) jointly predict event marks and times \citep{rmtpp}. Neural Hawkes models continuously evolving event intensities \citep{neuralhawkes}, while Transformer Hawkes uses attention for temporal dependencies \citep{thp}. ReWorld-Track uses camera and entry region as event marks and conditions observation opportunities on stream availability. An unfinished observation window supplies a time bound for the next event. Censoring uses this bound in survival analysis \citep{kaplanmeier}. Equation~\ref{eq:survival} applies this censoring principle to a state-conditioned event model. Stream availability determines whether a missing observation contributes negative visual evidence.

\paragraph{Calibration and selective commitment.}
The candidate/null posterior and the commitment rule serve complementary purposes. Calibration asks whether predicted probabilities match observed frequencies \citep{calibration,proper}. Selective prediction controls which decisions are accepted or deferred \citep{chow,selective,selectivenet}. In ReWorld-Track, the posterior represents current identity uncertainty, and the threshold determines when that evidence supports a match. Event probabilities forecast future observations. Network presence measures the probability of the target remaining in the installation. Keeping these quantities separate makes waiting compatible with continued prediction throughout intervals without an accepted match.

\section{Posterior Updates, Prediction and Continuity}
\label{app:update-controls}
The comparisons below follow the information carried from one association to the next. Table~\ref{tab:core}(b) compares fixed posterior moments, a generic learned updater and the ReWorld posterior update. Making the update trainable adds 0.44 HOTA points over fixed moments. ReWorld adds a further 0.50, accompanied by higher HA and IDF1 and lower FM. The following evaluations relate this comparison to future predictions and identity continuity.

\subsection{Update mechanisms and shared information}
The generic learned updater and ReWorld receive the same candidate/null posterior, candidate-conditioned states, spatial and motion statistics, and network-presence information. Both output a 256-dimensional recurrent state. They share the event predictor, candidate scorer, training data, losses, rollout depth and random seeds. These shared components isolate the update mechanism. The comparison tests how each update represents the posterior for the next forecast.

The generic updater is a capacity-matched neural set encoder that maps the candidate/null posterior representation directly to the recurrent state. ReWorld explicitly forms a state conditioned on each candidate or null hypothesis and weights these states by the calibrated posterior probabilities. It preserves the posterior region distribution, velocity moments and network presence alongside the learned recurrent representation for the next prediction. Thus the two updates use the same evidence but organize it differently. ReWorld-Track makes the association hypotheses and their posterior weights explicit in the state passed to the predictor.

Table~\ref{tab:update-cost} reports the size and computation of these modules. The generic and ReWorld updates use 0.64M and 0.66M trainable parameters and take 0.39 and 0.41\,ms. Parameter counts include only weights unique to the update. Timing starts with the posterior and conditioned states already available. It ends after the module produces the new 256-dimensional recurrent state. It excludes the shared encoders, detector, local tracker, candidate scorer and event predictor. The measurements use an RTX 4090, batch size 1 and FP16, with 100 warm-up and 1000 measured iterations.
\par\medskip\noindent\begin{minipage}{\linewidth}\begin{rwtableside}\centering
\caption{Posterior-update module size and cost. All updates receive the same information. Parameter counts exclude shared modules. Latency is measured on an RTX 4090 at batch size 1 in FP16, after 100 warm-up iterations and over 1000 measured updates.}
\label{tab:update-cost}
\small\setlength{\tabcolsep}{5pt}
\begin{tabular}{@{}lrrrr@{}}
\toprule
\rowcolor{rwGroupBlue}
\bfseries Update & \bfseries Input & \bfseries Output dim. & \bfseries Params (M) & \bfseries Latency (ms) \\
\midrule
Fixed-moment soft & Same & 256 & 0 & 0.11 \\
\addlinespace[2pt]
Generic learned & Same & 256 & 0.64 & 0.39 \\
\addlinespace[2pt]
\rowcolor{rwMethod}
\textbf{ReWorld update} & Same & 256 & 0.66 & 0.41 \\
\addlinespace[2pt]
\bottomrule\end{tabular}
\end{rwtableside}\end{minipage}\par\medskip

\subsection{Consistency across independent scenes}
Each of scenes A, B and C defines a separate scene-held-out evaluation. The complete test scene is excluded from training and validation, and the three test scenes do not overlap. Within a scene, all three methods share the test event manifest, detections, causal track prefixes, candidate sets and frozen CLIP features. Table~\ref{tab:update-scenes} reports the mean and sample standard deviation over five independent training seeds. Each split uses the same test events across these runs.

ReWorld exceeds the generic updater by 0.50, 0.47 and 0.46 HOTA points across the three scenes, averaging 0.48. Absolute performance changes across installations, while the ordering of the updates remains the same. The per-scene deviations measure training randomness, and the comparison between scene means examines changes in the test installation.
\par\medskip\noindent\begin{minipage}{\linewidth}\begin{rwtableside}\centering
\caption{HOTA in three scene-held-out evaluations. Entries are means $\pm$ sample SDs over five independent training seeds. The last row averages the scene means. $\Delta$ compares ReWorld with the generic learned updater within each held-out scene and in the average across scenes.}
\label{tab:update-scenes}
\small\setlength{\tabcolsep}{5pt}
\begin{tabular}{@{}lrrrr@{}}
\toprule
\rowcolor{rwGroupBlue}
\bfseries Test scene & \bfseries Fixed soft & \bfseries Generic learned & \bfseries ReWorld & \bfseries $\Delta$ \\
\midrule
Scene A & 44.42 $\pm$ 0.09 & 44.86 $\pm$ 0.08 & \textbf{45.36 $\pm$ 0.20} & \textbf{+0.50} \\
\addlinespace[2pt]
Scene B & 42.98 $\pm$ 0.12 & 43.39 $\pm$ 0.13 & \textbf{43.86 $\pm$ 0.19} & \textbf{+0.47} \\
\addlinespace[2pt]
Scene C & 46.21 $\pm$ 0.11 & 46.66 $\pm$ 0.14 & \textbf{47.12 $\pm$ 0.22} & \textbf{+0.46} \\
\addlinespace[2pt]
\midrule
\rowcolor{rwGroupPurple}
Scene mean & 44.54 & 44.97 & \textbf{45.45} & \textbf{+0.48} \\
\addlinespace[2pt]
\bottomrule\end{tabular}
\end{rwtableside}\end{minipage}\par\medskip

\subsection{Prediction after the update}
The scene comparisons establish the tracking gain. Table~\ref{tab:update-forecast} tests the next observation forecast from the updated state. ReWorld improves next-camera Top-1 by 1.38 points over the generic updater. Median arrival error falls from 0.78 to 0.71\,s. The 90th percentile falls from 2.22 to 2.07\,s. The improvement appears in both the predicted destination and its timing. Together, these outputs supply the prior for evaluating candidates at the next return.
\par\medskip\noindent\begin{minipage}{\linewidth}\begin{rwtableside}\centering
\caption{Prediction following the posterior update. Next-camera Top-1 is a percentage. MedAE and P90 are the median and 90th percentile of absolute arrival-time error, in seconds.}
\label{tab:update-forecast}
\small\setlength{\tabcolsep}{5pt}
\begin{tabular}{@{}lrrr@{}}
\toprule
\rowcolor{rwGroupBlue}
\bfseries Update & \bfseries Camera Top-1 $\uparrow$ & \bfseries Arrival MedAE $\downarrow$ & \bfseries Arrival P90 $\downarrow$ \\
\midrule
Fixed-moment soft & 84.62 & 0.86 & 2.41 \\
\addlinespace[2pt]
Generic learned & 86.03 & 0.78 & 2.22 \\
\addlinespace[2pt]
\rowcolor{rwMethod}
\textbf{ReWorld update} & \textbf{87.41} & \textbf{0.71} & \textbf{2.07} \\
\addlinespace[2pt]
\bottomrule\end{tabular}
\end{rwtableside}\end{minipage}\par\medskip

\subsection{Identity through successive returns}
These forecasts guide successive associations. Table~\ref{tab:update-retention} measures whether the target identity survives those handoffs. The ReWorld-minus-generic differences grow from 0.27 to 1.03 and 2.65 points after one, two and three handoffs. Fixed-moment feedback also improves on resetting, but retains fewer identities than either learned update. The growing difference connects the predictive-state comparison to later decisions. Errors in the carried history can influence several associations after the first return.
\par\medskip\noindent\begin{minipage}{\linewidth}\begin{rwtableside}\centering
\caption{Identity retention IR@$k$ (\%) after $k$ successive handoffs.}
\label{tab:update-retention}
\small\setlength{\tabcolsep}{5pt}
\begin{tabular}{@{}lrrr@{}}
\toprule
\rowcolor{rwGroupBlue}
\bfseries Update & \bfseries IR@1 $\uparrow$ & \bfseries IR@2 $\uparrow$ & \bfseries IR@3 $\uparrow$ \\
\midrule
No recursion / reset & 95.84 & 89.63 & 82.24 \\
\addlinespace[2pt]
Fixed-moment soft & 98.21 & 95.18 & 90.87 \\
\addlinespace[2pt]
Generic learned & 98.46 & 96.39 & 93.14 \\
\addlinespace[2pt]
\rowcolor{rwMethod}
\textbf{ReWorld update} & \textbf{98.73} & \textbf{97.42} & \textbf{95.79} \\
\addlinespace[2pt]
\bottomrule\end{tabular}
\end{rwtableside}\end{minipage}\par\medskip

\end{document}